\documentclass{article}

\usepackage{natbib}

\usepackage[utf8]{inputenc} % allow utf-8 input
\usepackage[T1]{fontenc}    % use 8-bit T1 fonts
\usepackage{hyperref}       % hyperlinks
\usepackage{url}            % simple URL typesetting
\usepackage{booktabs}       % professional-quality tables
\usepackage{amsfonts}       % blackboard math symbols
\usepackage{nicefrac}       % compact symbols for 1/2, etc.
\usepackage{microtype}      % microtypography
\usepackage{xcolor}         % colors
\usepackage{amsmath} % assumes amsmath package installed
\usepackage{amssymb}  % assumes amsmath package installed
\usepackage{xcolor}
\usepackage{algorithm}
\usepackage{algorithmic}
\usepackage{xspace}
\usepackage[small]{caption}
\usepackage{subcaption}
\usepackage{graphicx}
\usepackage{booktabs}
\usepackage{siunitx}
\usepackage{wrapfig}
\usepackage{amsthm}
\usepackage{mathtools}
\usepackage{thmtools,thm-restate}

\newcommand{\VI}[1]{{\color{black} #1}}
\newcommand{\MB}[1]{{\color{black} #1}}

\newtheorem{theorem}{Theorem}
\newtheorem{lemma}{Lemma}

\newtheorem{corollary}{Corollary}[theorem]

\title{Online POMDP Planning with Anytime Deterministic Optimality Guarantees}

\author{
  Moran Barenboim$^{1}$ \quad and \quad Vadim Indelman$^{2,3}$ \\
  \\
  $^{1}$ Technion Autonomous Systems Program (TASP) \\
  $^{2}$ Stephen B. Klein Faculty of Aerospace Engineering \\
  $^{3}$ Faculty of Data and Decision Sciences \\
  Technion - Israel Institute of Technology, Haifa 3200, Israel \\
}

\usepackage[nameinlink,noabbrev]{cleveref}
\begin{document}

\maketitle

%%%%%%%%%%%%%%%%%%%%%%%%%%%%%%%%%%%%%%%%%%%%%%%%%%%%%%%%%%%%%%%%%%%%%%%%%%%%%%%%
\begin{abstract}
Decision-making under uncertainty is a critical aspect of many practical
autonomous systems due to incomplete information. Partially Observable
Markov Decision Processes (POMDPs) offer a mathematically principled
framework for formulating decision-making problems under such conditions.
However, finding an optimal solution for a POMDP is generally intractable.
In recent years, there has been a significant progress of scaling
approximate solvers from small to moderately sized problems, using online
tree search solvers. Often, such approximate solvers are limited to
probabilistic or asymptotic guarantees towards the optimal solution. In this
paper, we derive a deterministic relationship for discrete POMDPs between an
approximated and the optimal solution. We show that at any time, we can
derive bounds that relate between the existing solution and the optimal one.
We show that our derivations provide an avenue for a new set of algorithms
and can be attached to existing algorithms that have a certain structure to
provide them with deterministic guarantees with marginal computational
overhead. In return, not only do we certify the solution quality, but we
demonstrate that making a decision based on the deterministic guarantee may
result in superior performance compared to the original algorithm without
the deterministic certification.

% Autonomous agents operating in real-world scenarios frequently encounter
% uncertainty and make decisions based on incomplete information. Planning
% under uncertainty can be mathematically formalized using partially
% observable Markov decision processes (POMDPs). However, finding an optimal
% plan for POMDPs can be computationally expensive and is feasible only for
% small tasks. In recent years, approximate algorithms, such as tree search
% and sample-based methodologies, have emerged as state-of-the-art POMDP
% solvers for larger problems. Despite their effectiveness, these algorithms
% offer only probabilistic and often asymptotic guarantees toward the optimal
% solution due to their dependence on sampling. To address these limitations,
% we derive a deterministic relationship between a simplified solution that is
% easier to obtain and the theoretically optimal one. First, we derive bounds
% for selecting a subset of the observations to branch from while computing a
% complete belief at each posterior node. Then, since a complete belief update
% may be computationally demanding, we extend the bounds to support reduction
% of both the state and the observation spaces. We demonstrate how our
% guarantees can be integrated with existing state-of-the-art solvers that
% sample a subset of states and observations. As a result, the returned
% solution holds deterministic bounds relative to the optimal policy. Lastly,
% we substantiate our findings with supporting experimental results.
\end{abstract}

%%%%%%%%%%%%%%%%%%%%%%%%%%%%%%%%%%%%%%%%%%%%%%%%%%%%%%%%%%%%%%%%%%%%%%%%%%%%%%%%

%%%%%%%%%%%%%%%%%%%%%%%%%%%%%
% Introduction %
%%%%%%%%%%%%%%%%%%%%%%%%%%%%%
\section{Introduction}
Decision-making under uncertainty is a common challenge in many practical
autonomous systems. In such systems, agents often operate with incomplete
information about their environment. This uncertainty can arise from various
sources, including sensor noise, hardware limitations, modeling approximations,
and the inherent unpredictability of the environment. Mathematically,
Decision-making under uncertainty can be formalized as Partially Observable
Markov Decision Process (POMDP). 

Unfortunately, finding an optimal solution to most POMDP problems is
computationally intractable, mostly due to a large number of possibilities for
the ground truth of the current state, and exponentially increasing
possibilities of the future outcomes, commonly referred to as the curse of
dimensionality, and the curse of history. As such, most state-of-the-art (SOTA)
algorithms aim to find an approximate solution.

One prominent approach to deriving approximate solutions employs an online
tree-search paradigm. In this framework, following each real-world decision, an
online solver evaluates the current state and projects potential future
scenarios. These scenarios are organized within a tree graph structure. As the
tree is constructed, the agent assesses the implications of selecting a
particular action, subsequently receiving feedback from the environment. This
feedback informs the estimation of probabilities for new states, guiding the
selection of subsequent actions based on accumulated knowledge. This iterative
process continues, building on past outcomes to navigate the decision space.

Given the inherent approximation in these solutions, a natural inquiry regarding
the connection between the approximate solution and the actual problem at hand.
Some state of the art online algorithms, e.g. \cite{Silver10nips}, offer
asymptotic guarantees thus having no finite time guarantees on the solution
quality. A different class of algorithms suggests finite time, but probabilistic
guarantees such as \cite{Somani13nips}. Many algorithms have shown good
empirical performance, at the advent of the practical use case of POMDP
problems, e.g. \cite{Sunberg18icaps}, but fall short of providing a framework
that bridges between the policy found and the underlying POMDP.

In this paper, we focus on deriving deterministic guarantees for POMDPs with
discrete state, action and observation spaces. Unlike existing black-box
sampling mechanisms employed in algorithms such as  \citep{Sunberg18icaps,
Hoerger21icra, Wu21nips}, our approach assumes access not only to the
observation model but also to the transition and the prior models. By leveraging
this additional information, we develop novel bounds that necessitate only a
subset of the state and observation spaces, enabling the computation of
deterministic bounds with respect to the optimal policy at any belief node
within the constructed tree. From a practical standpoint, we demonstrate how to
harness the theoretical derivations to recent advancements in POMDP approximate
solvers, by attaching the bounds to existing state-of-the-art algorithms. We
show that despite their stochastic nature, we can guarantee deterministic
linkage to the optimal solution with marginal computational overhead. We extend
the approach even further by demonstrating how to utilize the bounds to explore
the tree and finally select an action based on the deterministic guarantees.

In this paper, our main contributions are as follows. First, we introduce a
simplified POMDP that uses a subset of the state and observation spaces to
increase the computational efficiency. Then, we derive deterministic bounds that
relate between the former and the non-simplified POMDP. Notably, the bounds are
only a function of the states and observations known to the simplified POMDP and
hence can be calculated while planning to guide the decision-making and even
exploration. We also show a tighter version of the bounds considered in the
conference version of this paper, \citep{Barenboim2023neurips}. We further
extend the approach and show that utilizing these bounds for exploration results
in convergence to the optimal solution of the POMDP in finite time; While the
optimality guarantees applied only to observation-space simplification in the
conference version, we extend the results in this paper by deriving optimality
guarantees for both state- and observation-space simplification. Based on the
derived bounds, we illustrate how to incorporate the bounds into a general
structure of common state-of-the-art algorithms. We utilize the bounds for
exploration, decision-making and pruning of suboptimal actions while planning.
Last, we demonstrate the practicality of the bounds by experimenting with our
novel algorithms, suggested in this paper, namely DB-POMCP, RB-POMCP and
DB-DESPOT, which are variants of the POMCP and DESPOT algorithms, to improve the
empirical results in finite-horizon problems.

\begin{figure}[t]
	\centering
	\includegraphics[width=\textwidth]{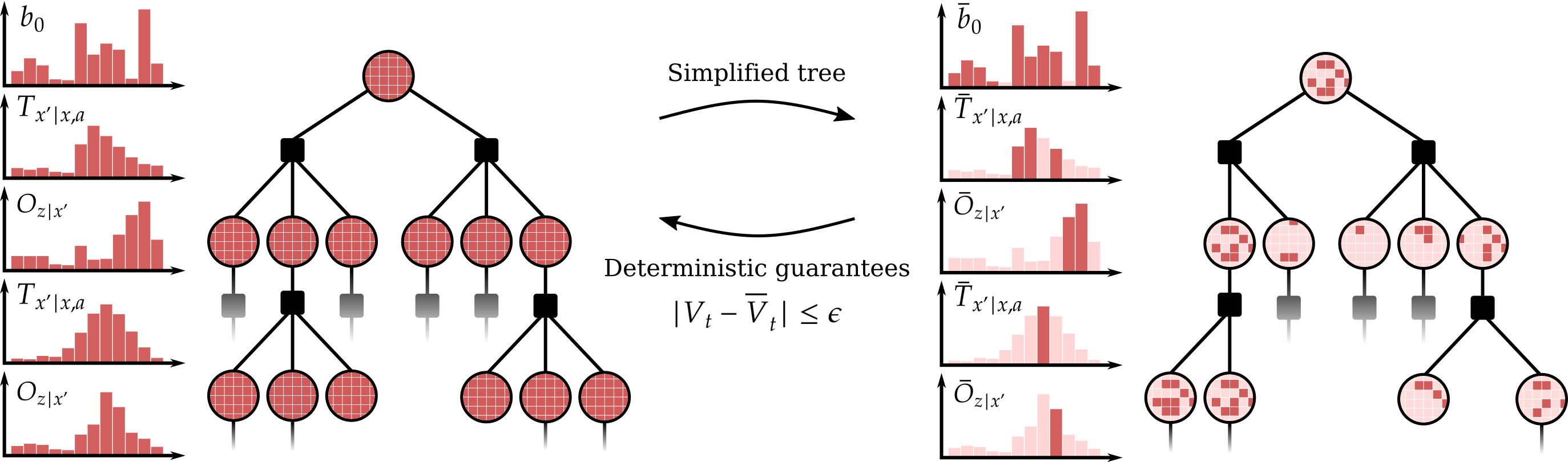}
	\label{fig:belief tree}
	\caption{The figure depicts two search trees: a complete tree (left) that considers all states and observations at each planning step, and a simplified tree (right) that incorporates only a subset of states and observations, linked to simplified models. Our methodology establishes a deterministic link between these two trees.}
\end{figure}

%%%%%%%%%%%%%%%%%%%%%%%%%%%%%
% Related Work %
%%%%%%%%%%%%%%%%%%%%%%%%%%%%%
\section{Related Work}
Over the last two decades there has been significant progress in online POMDP
planning, aiming to balance the trade-off between computational efficiency and
the quality of the solution.

The Heuristic Search Value Iteration (HSVI) \cite{Smith04uai} algorithm marked a
significant milestone in POMDP planning by introducing an efficient point-based
value iteration method that provides convergence guarantees. HSVI leverages a
heuristic to focus the search on the most promising regions of the belief space,
thus improving computational efficiency while maintaining solution quality.
Another pivotal algorithm, Successive Approximations of the Reachable Space
under Optimal Policies (SARSOP) \cite{Kurniawati08rss}, builds on this idea by
further refining the focus on reachable belief states under optimal policies.
SARSOP's ability to prune irrelevant parts of the belief space enables it to
handle larger POMDPs more effectively. However, these approaches were limited in
their scalability to large state spaces due to the necessity of computing a
complete belief state at each posterior node in the planning tree.

\MB{SARSOP, like other point-based offline solvers, provides suboptimality
guarantees \emph{after} completion of an offline solve, whereas our setting is
online decision making under tight per-decision budgets, yielding a certificate
at each iteration. In large spaces with concentrated posteriors, exhaustive
belief updates can allocate computation to low-probability regions;
sampling-based online planners (including ours) naturally bias effort toward
high-probability trajectories. Consequently, SARSOP is not our primary baseline
for time-constrained online planning. That said, its focus on “optimally
reachable” beliefs is complementary and can inform our selection of simplified
state/observation models without changing guarantees.}

The advent of Monte Carlo methods brought a significant shift in online POMDP
planning. The Partially Observable Monte Carlo Planning (POMCP)
\citep{Silver10nips} algorithm introduced a particle filter-based approach
combined with Monte Carlo tree search (MCTS). POMCP uses a set of particles to
represent the belief state and UCT (Upper Confidence bounds applied to Trees,
\citep{couetoux11book}) to guide the search, making it much more scalable and
for large state and observation spaces. POMCP is a forward search algorithm
which handles the large state and observation spaces by aggregating Monte-Carlo
rollouts of future scenarios in a tree structure. During each rollout, a single
state particle is recursively propagated from the root node to the leaves of the
tree. It adaptively trades off between actions that lead to unexplored areas of
the tree and actions that lead to rewarding areas of the tree search by
utilizing UCT \citep{Auer02ml}. The guarantees on the provided solution by POMCP
are asymptotic, implying that the quality of the solution remains unknown within
any finite time frame.

Another notable approximate solver, Anytime Regularized DESPOT (AR-DESPOT)
\citep{Somani13nips, Ye17jair} is derived from Regularized DESPOT, which holds
theoretical guarantees for the solution quality with respect to its optimal
value. Similar to POMCP, AR-DESPOT performs forward search and propagates a
single particle from the root node down to its leaves. It relies on
branch-and-bound approach in the forward search, and utilizes dynamic
programming techniques to update the value function estimate at each node. In
contrast to POMCP, Regularized DESPOT offers a probabilistic lower bound on the
value function obtained at the root node, providing a theoretical appeal by
measuring its proximity to the optimal policy.

While the primary focus of this paper is on discrete POMDP planning, it is
essential to acknowledge recent advancements in POMDP planning that encompass
both discrete and continuous observation spaces. Few notable approaches include
POMCPOW \citep{Sunberg18icaps}, LABECOP \citep{Hoerger21icra} and AdaOPS
\citep{Wu21nips}, which leverage explicit use of observation models. These
algorithms employ importance sampling mechanisms to weigh each state sample
based on its likelihood value, which is assumed to be known. Although these
methods have exhibited promising performance in practical scenarios, POMCPOW and
LABECOP currently lack formal guarantees, while \citep{Wu21nips} derives
probabilistic guarantees which do not hold in practice for AdaOPS algorithm, due
to assumption relaxations. To address this gap, \citep{Lim20ijcai, Lim23jair}
introduced a simplified solver aimed at bridging the theoretical gap between the
empirical success of these algorithms and the absence of theoretical guarantees
for continuous observation spaces. \VI{\citep{Lim23jair}, derive (i)
high-probability bounds on the value loss of their particle-belief solver and
(ii) a deterministic guarantee on its \emph{expected} return-obtained after
averaging over both the solvers internal randomness and the POMDPs stochastic
dynamics. Because this guarantee is only in expectation, a single execution may
still violate the bound, whereas the results presented in this paper hold
deterministically for every run.}

%%%%%%%%%%%%%%%%%%%%%%%%%%%%%
% Preliminaries %
%%%%%%%%%%%%%%%%%%%%%%%%%%%%%
\section{Preliminaries}
A finite horizon POMDP $M$ is defined as a tuple $\langle\mathcal{X},
\mathcal{A}, \mathcal{Z}, T, O, \mathcal{R}, b_0\rangle$, where $\mathcal{X}$,
$\mathcal{A}$, and $\mathcal{Z}$ represent a discrete state, action, and
observation spaces, respectively. The transition density function
$T(x_t,a_t,x_{t+1}) \triangleq \mathbb{P}(x_{t+1} | x_t, a_t)$ defines the
probability of transitioning from state $x_t \in \mathcal{X}$ to state $x_{t+1}
\in \mathcal{X}$ by taking action $a_t \in \mathcal{A}$. The observation density
function $O(x_t,z_t) \triangleq \mathbb{P}(z_t | x_t)$ expresses the probability
of receiving observation $z_t \in \mathcal{Z}$ from state $x_t \in \mathcal{X}$.
$b_0\equiv \mathbb{P}(x_0 \mid H_0)$ represents the prior probability function,
which is the distribution function over the state space at time $t=0$.

Given the limited information provided by observations, the true state of the
agent is uncertain and a probability distribution function over the state space,
also known as a belief, is maintained. The belief depends on the entire history
of actions and observations, denoted as $H_t \triangleq \{z_{1:t}, a_{0:t-1}\}$.
We also define the propagated history as $H_t^- \triangleq \{z_{1:t-1},
a_{0:t-1}\}$. At each time step $t$, the belief is updated by applying Bayes'
rule using the transition and observation models, given the previous action
$a_{t-1}$ and the current observation $z_t$, $b \left( x_t \right) = \eta_t
\mathbb{P}(z_t | x_t ) \sum_{x_{t-1} \in \mathcal{X}}  
 \mathbb{P}(x_t | x_{t-1}, a_{t-1} ) b \left( x_{t-1}
\right) $, where $\eta_t$ denotes a normalization constant and
$b_t\triangleq\mathbb{P}(x_t\mid H_t)$ denotes the belief at time t. The updated
belief, $b_t$, is  sometimes referred to as the posterior belief, or simply the
posterior. We will use these interchangeably throughout the paper.

A policy function $a_t = \pi_t(H_t)$ determines the action to be taken at time
step $t$, based on the history $H_t$ and time $t$. In the rest of the paper we
write $\pi_t \equiv \pi_t(H_t)$ for conciseness. The reward is defined as an
expectation over a state-dependent function, $r(b_t, a_t)=\mathbb{E}_{x\sim
b_t}[r_x(x, a_t)]$, and is assumed to be bounded by $-\mathcal{R}_{\max}\leq
r_x(x,a_t)\leq \mathcal{R}_{\max}$. The value function for a policy $\pi$ over a
finite horizon $T$ is defined as the expected cumulative reward
received by executing $\pi$ and can be computed using the Bellman update
equation,
\begin{equation} \label{eq: bellman update}
	V^{\pi }_t( b_{t}) =r( b_{t } ,\pi_t) + \underset{z_{t+1:T}}{\mathbb{E}} \left[ \sum _{\tau =t+1}^{T}r( b_{\tau } ,\pi _\tau)\right].
\end{equation}
We use $V^{\pi }_t( b_{t})$ and $V^{\pi }_t( H_{t})$ interchangeably throughout
the paper. The action-value function is defined by executing action $a_t$ and
then following policy $\pi$,
\begin{equation}
	Q^\pi_t( b_{t}, a_t) =r( b_{t } ,a_t) + \underset{z_{t+1:T}}{\mathbb{E}} \left[ \sum _{\tau =t+1}^{T}r( b_{\tau } ,\pi _\tau)\right].
\end{equation}
The optimal value function may be computed using
Bellman's principle of optimality, %\VI{[should be $\pi^*$?]}
\begin{equation} \label{eq: bellman optimality}
	V^{\pi^* }_t( b_{t}) =\max_{a_t}\{r( b_{t } , a_t) + \underset{z_{t+1}\mid a_t, b_t}{\mathbb{E}} \left[ V^{\pi^* }_{t+1}( b_{t+1})\right]\}.
\end{equation}

The goal of the agent is to find the optimal policy $\pi^*$ that maximizes the
value function.

For notational convenience, we introduce a few more simplifying notations; We
use $\mathcal{V}_{max,t}, \mathcal{V}_{min,t}$ to denote upper an lower bounds
on the value function at time step $t$. In the simplest case, these may be
$\mathcal{V}_{max,t}=(T-t)\cdot \mathcal{R}_{\max}$,
$\mathcal{V}_{min,t}=(t-T)\cdot \mathcal{R}_{\max}$. Additionally, we denote a
trajectory as, $\tau_t=\{x_0,a_0,z_1,x_1,a_1,\dots, a_{t-1}, x_t, z_t\}$, and a
corresponding probability distribution over the possible trajectories,
$\mathbb{P}(\tau_t)$. We denote a policy-dependent trajectory distribution as
$\mathbb{P}^\pi(\tau_t)\equiv \mathbb{P}(\tau_t\mid b_0, \pi_0, \dots, \pi_t)$.

%%%%%%%%%%%%%%%%%%%%%%%%%%%%%
% Methods %
%%%%%%%%%%%%%%%%%%%%%%%%%%%%%
\section{Simplified POMDP} \label{sec:methods}
Typically, it is infeasible to fully expand a Partially Observable Markov
Decision Process (POMDP) tree due to the extensive computational resources and
time required. To address this challenge, we propose two approaches. In the
first approach, presented in \ref{subsec:observationSimplify}, we propose a
solver that selectively chooses a subset of the observations to branch from,
while maintaining a full posterior belief at each node. This allows us to derive
an hypothetical algorithm that directly uses our suggested deterministic bounds
to choose which actions to take while exploring the tree. As in most scenarios
computing a complete posterior belief may be too expensive, in section
\ref{subsec:StateObsSimplify} we suggest an improved method that in addition to
branching only a subset of the observations, selectively chooses a subset of the
states at each encountered belief. 

The presented approaches diverge from many existing algorithms that rely on
black-box prior, transition, and observation models. Instead, our method
directly utilizes state and observation probability values to evaluate both the
value function and the associated bounds. In return, an anytime deterministic
guarantee on the value function for the derived policy concerning its deviation
from the optimal value function is derived.

To that end, we define a simplified POMDP, which is a reduced version of the
original POMDP that abstracts or ignores certain states and/or observations. A
simplified POMDP, $\bar{M}$, is a tuple $\langle\bar{\mathcal{X}}, \mathcal{A},
\bar{\mathcal{Z}}, \bar{T}, \bar{O}, \mathcal{R}, \bar{b}_0\rangle$, where
$\bar{\mathcal{X}}, \bar{\mathcal{Z}}, \bar{T}$ and $\bar{O}$ are the simplified
versions of the state and observation spaces, and their corresponding transition
and observation models,
\begin{align}
    \bar{b}_0(x) \triangleq& \begin{cases}
        b_0(x) & ,  \ x \in \bar{\mathcal{X}}_0 \label{eq:simplifiedPrior}\\
        0 & , \ otherwise
    \end{cases}\\
    \bar{\mathbb{P}}(x_{t+1}\mid x_t, a_t) \triangleq& \begin{cases}
        \mathbb{P}(x_{t+1}\mid x_t,
a_t) & ,  \ x_{t+1} \in \bar{\mathcal{X}}(H_{t+1}^-) \label{eq:simplifiedTransition}\\
        0 & , \ otherwise
        \end{cases}\\
        \bar{\mathbb{P}}(z_{t}\mid x_t) \triangleq& \begin{cases}
            \mathbb{P}(z_{t}\mid x_t) & ,  \ z_t \in \bar{\mathcal{Z}}(H_t)\label{eq:simplifiedObservation}\\
            0 & , \ otherwise
        \end{cases}
\end{align}
where $\bar{\mathcal{\mathcal{X}}}(H_{t+1}^-)\subseteq \mathcal{\mathcal{X}}$
and $\bar{\mathcal{Z}}(H_t)\subseteq \mathcal{Z}$ may be chosen arbitrarily,
e.g. by sampling or choosing a fixed subset a-priori, as the derivations of the
bounds are independent of the subset choice. Note that the simplified prior,
transition and observation models are unnormalized and do not aim to represent
valid distribution functions. For the rest of the sequel we drop the explicit
dependence on the history, and denote
$\bar{\mathcal{X}}(H_{t+1}^-)\equiv\bar{\mathcal{X}}$,
$\bar{\mathcal{Z}}(H_t)\equiv\bar{\mathcal{Z}}$. The action space, $\mathcal{A}$
and prior probability, $b_0$ are as defined in the original POMDP, $M$.

With the definition of the simplified POMDP, we define a corresponding
simplified value function,
\begin{align} \label{def:simplifiedValueFunc}
    \bar{V}^\pi(\bar{b}_0)&\triangleq \bar{\mathbb{E}}\left[\sum_{t=0}^{T}r(x_t, a_t)\right]\\
    &=\sum _{t=0}^{T} \sum _{z_{1:t}}\sum _{x_{0:t}}\prod _{k=1}^{t} \ \bar{\mathbb{P}}( z_{k} \mid x_{k})\bar{\mathbb{P}}( x_{k} \mid x_{k-1} ,\pi _{k-1}) \bar{b}( x_{0}) r( x_{t} ,a_{t})\\
    &= \sum_{t=0}^T \sum_{\tau_{t}} \bar{\mathbb{P}}^\pi(\tau_t) r(x_t, a_t),
    % &= \sum_{t=0}^T \sum_{\tau_{t}} \bar{\mathbb{P}}^\pi(\tau_t) \left[\sum_{\tau_{T}}\mathbb{P}^\pi(\tau_T\mid \tau_t)\right] r(x_t, a_t) \\
    % &= \sum_{t=0}^T \sum_{\tau_{T}} \mathbb{P}^\pi(\tau_T\mid \tau_t) \bar{\mathbb{P}}^\pi(\tau_t) r(x_t, a_t) \\
    % &= \sum_{\tau_{T}} \mathbb{P}^\pi(\tau_T\mid \tau_t) \bar{\mathbb{P}}^\pi(\tau_t) \sum_{t=0}^T  r(x_t, a_t) \\
    % &= \sum_{\tau_{T}} \mathbb{P}^\pi(\tau_T\mid \tau_t) \bar{\mathbb{P}}^\pi(\tau_t) v(\tau_T)
\end{align}
where the simplified expectation-like operator, $\bar{\mathbb{E}}[\cdot]$, is
taken with respect to the simplified prior, transition and observation models,
which do not include the entire distribution, and thus is not a complete
expectation.

We use the simplified value function as a computationally-efficient replacement
for the theoretical value function; For clarity, the simplified POMDP and
consequently all derivations consider a finite-horizon POMDP, but its extension
to the discounted infinite horizon case is straightforward, by introducing the
discount factor whenever the reward is being used, and an additive term for
truncating the tree, $\gamma^t V_{max,t}$, as suggested in, e.g.,
\cite{Kocsis06ecml}.

In the following sections, we will derive upper and lower bounds between the
simplified and the theoretical values of a given policy. Then, we will show how
to use the simplification to achieve guarantees with respect to the optimal
value function of the original POMDP, and how to utilize these bounds for
planning.

%%%%%%%%%%%%%%%%%%%%%%%%%%%%%
% Algorithms %
%%%%%%%%%%%%%%%%%%%%%%%%%%%%%
\section{Anytime Deterministic Guarantees for Simplified POMDPs} \label{sec:analysis}

\subsection[short]{Simplified Observation Space}
\label{subsec:observationSimplify}
We first analyze the performance guarantees of a simplified observation space,
while assuming a complete belief update at each considered history node, i.e.,
$\bar{\mathcal{X}}\equiv \mathcal{X}$. Such an approach is viable when the
posterior belief can be calculated efficiently, e.g. when the state spae is
sufficiently small. We start by presenting a bound between the simplified value
function and the theoretical one of a given policy; then, we provide optimality
guarantees for any policy, obtained by solving the simplified POMDP, both in
terms of convergence and a deterministic bound, in which the optimal value, for
an unknown policy must reside in.

\subsubsection{Fixed Policy Guarantees for Simplified Observation Spaces}
 The following theorem describes the guarantees of the
observation-simplified value function with respect to its theoretical value,

\begin{theorem} \label{thm:valueFuncSimplifiedObs}
    Let $b_t$ belief state at time $t$, and $T$ be the last time step of the POMDP.
    Let $V^{\pi}(b_t)$ be the theoretical value function by following a policy
    $\pi$, and let $\bar{V}^{\pi}(b_t)$ be the simplified value function, 
    as defined in \eqref{def:simplifiedValueFunc}, by following the same policy.
    Then, for any policy $\pi$, the difference between the theoretical and
    simplified value functions is bounded as follows,
    \begin{equation} \label{eq:thm}
        \left|V^\pi(b_t) \! - \! \bar{V}^\pi(b_t)\right|\leq \! \mathcal{R}_{\max} \! \! \! \sum _{\tau=t+1}^{T} \!\! \left[ 1\! -\! \! \! \sum_{z_{t+1:{\tau}}}\sum
        _{x_{t:{\tau}}}b( x_{t}) \! \! \prod _{k=t+1}^{\tau}\overline{\mathbb{P}}( z_{k} \mid x_{k})
        \mathbb{P}( x_{k}\mid x_{k-1} ,\pi _{k-1})\right]\triangleq \epsilon^\pi(b_t).
    \end{equation}
\end{theorem}
\begin{proof}
    The proof is provided in \Cref{app:proof-thm-valueFuncSimplifiedObs}.
\end{proof}

Similarly, the action-dependent bound on the value difference, denoted $\epsilon^\pi(b_t,
a_t)$, is the bound of taking action $a_t$ in belief $b_t$ and following policy
$\pi$ thereafter,
\begin{equation}
    \left|Q^\pi(b_t, a_t) \! - \! \bar{Q}^\pi(b_t, a_t)\right| \leq \epsilon^\pi(b_t, a_t),
\end{equation}
where,
\begin{align} \label{eq:Q_z_bound}
    \epsilon^\pi(b_t, a_t) \triangleq 
    \mathcal{R}_{\max}\sum _{\tau=t+1}^{T}\big[ 1-\sum_{z_{t+1:{\tau}}}\sum
    _{x_{t:{\tau}}}b( x_{t})\overline{\mathbb{P}}( z_{t+1} \mid x_{t+1})
    \mathbb{P}( x_{t+1}\mid x_{t} ,a_{t})\cdot& \\
    \prod _{k=t+2}^{\tau}\overline{\mathbb{P}}( z_{k} \mid x_{k})
    \mathbb{P}( x_{k}\mid x_{k-1} ,\pi _{k-1})\big].&\notag
\end{align}

Importantly, $\epsilon^\pi(b_t)$ and $\epsilon^\pi(b_t, a_t)$ only contain terms
which depend on observations that are within the simplified space, $z\in
\bar{\mathcal{Z}}$. This is an essential property of the bounds, as it is a
value that can easily be calculated during the planning process and provides a
certification of the policy quality at any given node along the tree.
Furthermore, it is apparent from \eqref{eq:thm} that as the number of
observations included in the simplified set, $\bar{\mathcal{Z}}$, increases, the
values of $\epsilon^\pi(b_t)$ and $\epsilon^\pi(b_t, a_t)$ consequently
diminishes, 
\begin{equation}
    \sum_{z_{1:{\tau}}}\sum
_{x_{0:{\tau}}}b( x_{0})\prod _{k=1}^{\tau}\overline{\mathbb{P}}( z_{k} \mid x_{k})
\mathbb{P}( x_{k}\mid x_{k-1} ,\pi _{k-1}) \xrightarrow{\bar{\mathcal{Z}}\rightarrow \mathcal{Z}} 1 \notag
\end{equation}
leading to a convergence towards the theoretical value function, i.e.
$\epsilon^\pi(b_t)\rightarrow 0$ and $\epsilon^\pi(b_t, a_t)\rightarrow 0$.

\subsubsection{Optimality Guarantees for Simplified Observation Spaces} 
Theorem \ref{thm:valueFuncSimplifiedObs} provides both lower and upper bounds
for the theoretical value function, assuming a fixed policy. Using this theorem,
we can derive upper and lower bounds for any policy, including the optimal one.
This is achieved by applying the Bellman optimality operator to the upper bound
in a repeated manner, instead of the estimated value function; In the context of
tree search algorithms, our algorithm explores only a subset of the decision
tree due to pruned observations. However, at every belief node encountered
during this exploration, all potential actions are expanded.  The action-value
function of these expanded actions is bounded using the Upper Deterministic
Bound, which we now define as
\begin{equation}
    \textsc{UDB}^{\pi}(b_t,a_t) \triangleq \bar{Q}^{\pi}(b_t, a_t) + \epsilon^\pi(b_t, a_t) = r(b_t, a_t) + \bar{\mathbb{E}}_{z_{t+1}}[\bar{V}^\pi(b_{t+1})] + \epsilon^\pi(b_t, a_t).
\end{equation}

In the event that no subsequent observations are chosen for a given history, the
value of $\bar{Q}^{\pi}(b_t, a_t)$ simplifies to the immediate reward plus an
upper bound for any subsequent policy, given by $\mathcal{R}_{\max}\cdot
(T-t-1)$. Then, we make the following claim,

\begin{lemma}\label{lemm1} The optimal value function can be bounded by,
	\begin{equation}
		V^{\pi*}(b_t)  \leq    \textsc{UDB}^{\pi^{\dagger}}(b_t),
	\end{equation}
	where the policy $\pi^{\dagger}$ is determined according to Bellman
	optimality over the UDB, i.e.
	\begin{gather}
            \pi^{\dagger}(b_t) = \arg \max_{a_t \in \mathcal{A}} [\bar{Q}^{\pi^{\dagger}}(b_t, a_t) + \epsilon^{\pi^{\dagger}}(b_t, a_t)] = \arg \max_{a_t \in \mathcal{A}} \textsc{UDB}^{\pi^{\dagger}}(b_t,a_t)\\
		\textsc{UDB}^{\pi^{\dagger}}(b_t) \triangleq  \max_{a_t \in \mathcal{A}} \textsc{UDB}^{\pi^{\dagger}}(b_t,a_t).
	\end{gather}
	
\end{lemma}
\begin{proof}
    The proof is provided in \Cref{app:proof-lemm1}.
\end{proof}

Notably, using UDB to find the optimal policy does not require a recovery of all
the observations in the theoretical belief tree, but only a subset which depends
on the definition and complexity of the POMDP. Each action-value is bounded by a
lower and upper bound, which can be represented as an interval enclosing the
theoretical value. When the bound intervals of two candidate actions do not
overlap, one can clearly discern which action is suboptimal, rendering its
subtree redundant for further exploration. This distinction sets UDB apart from
current state-of-the-art online POMDP algorithms. In those methods, any
finite-time stopping condition fails to ensure optimality since the bounds used
are either heuristic or probabilistic in nature. \VI{Note, however, that
previous \emph{offline} algorithms did utilize similar pruning, such as SARSOP.}

In addition to certifying the obtained policy with Bellman optimality criteria,
one can utilize UDB as an exploration criteria,
\begin{equation}\label{eq:exploration}
	a_t = \arg \max_{a_t \in \mathcal{A}}[   \textsc{UDB}^{\pi^{\dagger}}(b_t,a_t)],
\end{equation}
which ensures convergence to the optimal value function, as the number of
visited posterior nodes increases.
\begin{corollary} \label{cor:1}
	By utilizing Lemma \ref{lemm1} and the exploration criteria defined in
	\eqref{eq:exploration}, an increasing number of explored belief nodes
	guarantees convergence to the optimal value function.
\end{corollary}
\begin{proof}
    The proof is provided in \Cref{proof:corollary1.1}.
\end{proof}

\subsection[short]{Simplified State and Observation Spaces}
\label{subsec:StateObsSimplify}
In most scenarios, a complete evaluation of posterior beliefs during the
planning stage may pose significant computational challenges. To tackle this
issue, we propose the use of a  simplified state space in addition to the
simplified observation space considered thus far. Specifically, we derive
deterministic guarantees of the value function that allow for the selection of a
subset from both the states and observations.

We start the analysis of simplifying the state-and-observation spaces by fixing
a policy and derive upper and lower bounds for the theoretical, yet unknown,
value function at the root node, hereafter referred to as the 'root-value'. This
process involves the use of a simplified value function and an additional bonus
term, which are easier to compute than the theoretical value function.
Considering that various segments of the decision tree contribute differently to
the upper bound, we then examine each subtree's contribution separately, which
leads to a recursive formulation of the bound. Importantly, these bounds are
exclusively derived in relation to, and hold only with respect to, the root node. This is in contrast to the
bounds shown in theorem \ref{thm:valueFuncSimplifiedObs}, which bound the value
function of each node in the belief tree.

Using the deterministic bounds at the root allows us to certify the performance
of following a particular policy starting from the root of the planning tree.
Based on these bounds we extend previous results, shown in theorem
\ref{thm:valueFuncSimplifiedBoth}, and show that, (1) exploring the tree with a
bound that is formulated with respect to the root node leads to an optimistic
estimation of the optimal value function with respect to that root node. (2)
Utilizing the bounds for action exploration leads to convergence to the optimal
solution of the entire tree.
% (3) We develop a new method for pruning suboptimal
% mid-tree action branches. This method includes a bonus term for the upper and
% lower bounds, accounting for previously unconsidered cumulative probability,
% enhancing model efficiency by eliminating less optimal paths.

\subsubsection{Fixed policy guarantees}
We begin by stating the core theorem of our paper, which sets forth the upper
and lower bounds of a root-value function, with a simplified value function, 

\begin{theorem} \label{thm:valueFuncSimplifiedBoth}

    Let $b_0$ and $\bar{b}_0$ be the theoretical and simplified belief states,
    respectively, at time $t=0$, and $T$ be the last time step of the POMDP. Let
    $V^{\pi}(b_0)$ be the theoretical value function by following a policy
    $\pi$, and let $\bar{V}^{\pi}(\bar{b}_0)$ be the simplified value function
    by following the same policy, as defined in \eqref{def:simplifiedValueFunc}.
    Then, for any policy $\pi$, the theoretical value function and at the root
    is bounded as follows, 
    \begin{equation}
        \mathcal{L}^\pi_0(H_0) \leq V^\pi(b_0) \leq \mathcal{U}^\pi_0(H_0).
    \end{equation}
    where, 
    \begin{align}
        \mathcal{U}^\pi_0(H_0) &\equiv \bar{V}^\pi(\bar{b}_0) + \mathcal{V}_{max,0}\left[1-\sum_{\tau_{0}}\bar{\mathbb{P}}(\tau_{0})\right]+\sum _{t=0}^{T-1}\mathcal{V}_{max,t+1}\left[ \sum_{\tau_{t}} \bar{\mathbb{P}}^{\pi}(\tau_{t})-\sum_{\tau_{t+1}}\bar{\mathbb{P}}^{\pi}(\tau_{t+1})\right] \label{eq:upperBound}\\ 
        \mathcal{L}^\pi_0(H_0) &\equiv \bar{V}^\pi(\bar{b}_0) + \mathcal{V}_{min,0}\left[1-\sum_{\tau_{0}}\bar{\mathbb{P}}(\tau_{0})\right]+\sum _{t=0}^{T-1}\mathcal{V}_{min,t+1}\left[ \sum_{\tau_{t}} \bar{\mathbb{P}}^{\pi}(\tau_{t})-\sum_{\tau_{t+1}}\bar{\mathbb{P}}^{\pi}(\tau_{t+1})\right] \label{eq:lowerBound}
    \end{align}
    \begin{proof} The proof is provided in \Cref{proof:thm2}.
    \end{proof}
\end{theorem}
In this theorem, we introduced a minor modification to the theorem presented in
the conference version of this paper, \cite{Barenboim2023neurips}. We replaced
the term $\mathcal{R}_{\max}\cdot (T-t)$ with the more general
$\mathcal{V}_{max,t}$
by
performing simple algebraic transitions; the principles and conclusions of both
remain the same. A key aspect of Theorem \ref{thm:valueFuncSimplifiedBoth} is that the bounds it establishes
are exclusively dependent on the simplified state and observation spaces. This
characteristic is vital in order to compute them during the planning phase.

The intuition behind the result of the derivation can be interpreted as follows;
it takes a conservative approach to the value estimation by assuming that every
trajectory not observed may obtain an extremum value. Moreover, it allows
flexibility in how the trajectories are selected, which are allowed to be chosen arbitrarily
in terms of the simplified state space, observation space and the horizon of
each trajectory. 

The theorem provides bounds for the theoretical value function at the
root node of the search tree, given a policy. Using Bellman-like equations, one can restructure the
formulation to compute the bounds recursively, which is crucial for making
computations in online planning computationally efficient,
\begin{align} \label{def:RecursiveBound}
    % &\mathcal{U}^\pi_0(H_t) \triangleq \sum_{\tau_{t}} \bar{\mathbb{P}}(\tau_t \mid H_t) r(x_t, \pi_t) + \sum_{z_{t+1}\in \bar{\mathcal{Z}}(H^-_{t+1})}\left[\mathcal{U}^\pi_0(H_{t+1}) + \mathcal{V}_{\max, t} \left(\sum_{\tau_t}\bar{\mathbb{P}}(\tau_t \mid H_t) - \sum_{\tau_{t+1}}\bar{\mathbb{P}}(\tau_{t+1} \mid H_{t+1})\right)\right]\\
    % &\mathcal{L}^\pi_0(H_t) \triangleq \sum_{\tau_{t}} \bar{\mathbb{P}}(\tau_t \mid H_t) r(x_t, \pi_t) + \sum_{z_{t+1}\in \bar{\mathcal{Z}}(H^-_{t+1})}\left[\mathcal{L}^\pi_0(H_{t+1}) + \mathcal{V}_{\min, t} \left(\sum_{\tau_t}\bar{\mathbb{P}}(\tau_t \mid H_t) - \sum_{\tau_{t+1}}\bar{\mathbb{P}}(\tau_{t+1} \mid H_{t+1})\right)\right]\\
    \mathcal{U}^\pi_0(H_t) \triangleq &\sum_{\tau_{t}\in \mathcal{T}(H_t)} \bar{\mathbb{P}}(\tau_t) r(x_t, \pi_t) + \sum_{\tau_t \in \mathcal{T}(H_t)}\bar{\mathbb{P}}(\tau_t)\mathcal{V}_{\max, t} \\ \notag 
    &+ \sum_{z_{t+1}\in \bar{\mathcal{Z}}(H_t, \pi_t)}\left[\mathcal{U}^\pi_0(H_{t+1}) - \sum_{\tau_{t+1}\in \mathcal{T}(H_{t+1})}\!\!\!\!\!\!\!\!\!\bar{\mathbb{P}}(\tau_{t+1})\mathcal{V}_{\max, t}\right]\\
    \mathcal{L}^\pi_0(H_t) \triangleq &\sum_{\tau_{t}\in \mathcal{T}(H_t)} \bar{\mathbb{P}}(\tau_t) r(x_t, \pi_t) + \sum_{\tau_t \in \mathcal{T}(H_t)}\bar{\mathbb{P}}(\tau_t)\mathcal{V}_{\min, t}  \\ \notag  
    &+  \sum_{z_{t+1}\in \bar{\mathcal{Z}}(H_t, \pi_t)}\left[\mathcal{L}^\pi_0(H_{t+1}) - \sum_{\tau_{t+1}\in \mathcal{T}(H_{t+1})}\!\!\!\!\!\!\!\!\!\bar{\mathbb{P}}(\tau_{t+1})\mathcal{V}_{\min, t}\right]
\end{align}
and,
\begin{align}
    &\mathcal{U}^\pi_0(H_T) \triangleq \sum_{\tau_{T}\in\mathcal{T}(H_T)} \bar{\mathbb{P}}(\tau_T) r(x_T),
    &\mathcal{L}^\pi_0(H_T) \triangleq \sum_{\tau_{T}\in\mathcal{T}(H_T)} \bar{\mathbb{P}}(\tau_T) r(x_T).
\end{align}
where $\mathcal{T}(H_t)$ represent the set of trajectories that consist history
$H_t$, i.e., all trajectories $\mathcal{T}(H_t)=\{\left(x_{0:t}, a_{0:t-1},
z_{1:t}\right)\mid (a_{0:t-1}, z_{1:t})=H_t\}$. The values $\mathcal{U}^\pi_0(H_t)$
and $\mathcal{L}_0^\pi(H_t)$, represent the relative upper and lower bounds of
node $H_t$ with respect to the value function at the root, $H_0$. In other
words, they do not represent the bounds of a policy starting from node $H_t$.
The first two summands have a similar structure to the standard Bellman update
operator used in POMDPs, with two main differences. First, the state dependent
reward is multiplied by the probability of the entire trajectory from the root
node, and not the density value of the belief. Notably, the value of
$\sum_{\tau_{t}\in \mathcal{T}(H_t)} \mathbb{P}(\tau_t)$ will generally not sum to one, due
to the dependence of the summed trajectories on the history. Second, there is no
expectation operator over the values of the next time step. This is a result of
using a distribution over the trajectories, instead of the belief itself. The
last summand assigns an optimistic value for the set of trajectories reached to
node $H_t$ but not to $H_{t+1}$. 

\subsubsection{Optimality Guarantees}
We have shown in theorem \ref{thm:valueFuncSimplifiedBoth} how to calculate
bounds for the difference in value functions between the original and the simplified POMDP, given a fixed
policy. In this section, we show
that by applying Bellman-like optimality operator on $\mathcal{U}_0(H_t)$, the
obtained value at the root node is an upper bound for the optimal value
function. More formally,
\begin{corollary}\label{lemma:udbOptimality} Let $\mathcal{A}$ be the set of
    actions and $\mathcal{U}_0^\star(H_t)$, $\mathcal{L}_0^\star(H_t)$ be the upper
    and lower bounds of node $H_t$ chosen according to,
    \begin{align} 
        \mathcal{U}^\star_0(H_t) &\triangleq \max_{a_t\in\mathcal{A}} \sum_{\tau_{t}\in \mathcal{T}(H_t)} \bar{\mathbb{P}}(\tau_t) \left[r(x_t, a_t) + \mathcal{V}_{\max, t}\right]  \label{eq:optimalRecursive}\\
        &+ \sum_{z_{t+1}\in \bar{\mathcal{Z}}(H_t, a_t)}\left[\mathcal{U}^\star_0(H_{t+1}) - \sum_{\tau_{t+1}\in \mathcal{T}(H_{t+1})}\!\!\!\!\!\!\!\!\!\bar{\mathbb{P}}(\tau_{t+1})\mathcal{V}_{\max, t}\right] \notag \\
        \mathcal{L}^\star_0(H_t) &\triangleq \max_{a_t\in\mathcal{A}} \sum_{\tau_{t}\in \mathcal{T}(H_t)} \bar{\mathbb{P}}(\tau_t) \left[r(x_t, a_t) + \mathcal{V}_{\min, t}\right]   \label{eq:optimalRecursive2}\\
        &+ \sum_{z_{t+1}\in \bar{\mathcal{Z}}(H_t, a_t)}\left[\mathcal{L}^\star_0(H_{t+1}) - \sum_{\tau_{t+1}\in \mathcal{T}(H_{t+1})}\!\!\!\!\!\!\!\!\!\bar{\mathbb{P}}(\tau_{t+1})\mathcal{V}_{\min, t}\right] \notag
    \end{align}
    and,
    \begin{align}
        &\mathcal{U}_0^\star(H_T) \triangleq \sum_{\tau_{T}\in \mathcal{T}(H_T)} \bar{\mathbb{P}}(\tau_T) r(x_T),
        &\mathcal{L}_0^\star(H_T) \triangleq \sum_{\tau_{T}\in \mathcal{T}(H_T)} \bar{\mathbb{P}}(\tau_T) r(x_T).
    \end{align}

    Then, the optimal root-value is bounded by,
    \begin{equation}
    \mathcal{L}_0^\star(H_0) \leq V^{\pi^*}(H_0) \leq \mathcal{U}_0^\star(H_0).
    \end{equation}
    \begin{proof} 
    The proof is provided in \ref{proof:optimalRecursive}.
    \end{proof}
\end{corollary}
In this corollary, we establish that employing the 'partial' root-bound is
sufficient for ensuring both upper and lower bounds in relation to the optimal
value function at the root node. This approach differs from that presented in
the previous section (see Lemma \ref{lemm1}). There, each node in the tree was
associated with its unique upper bound based on its value function. In contrast,
the current corollary demonstrates that using the 'partial' bound across all nodes
in the tree, which is valid only at the root, still guarantees bounded value for
the optimal root-value function, while avoiding the requirement to maintain a
complete belief at each node of the tree.

\subsubsection{Early Stopping Criteria} \label{subsec:pruningStoppingCriteria}

Corollary \ref{lemma:udbOptimality} establishes that the recursive Bellman-like
optimality operator, can be used to bound the optimal value function at the
root. Since the bounds are deterministic, these bounds can be used for
eliminating suboptimal actions with full certainty while planning.

\begin{figure} [h]
	\centering
	\includegraphics[width=.5\linewidth]{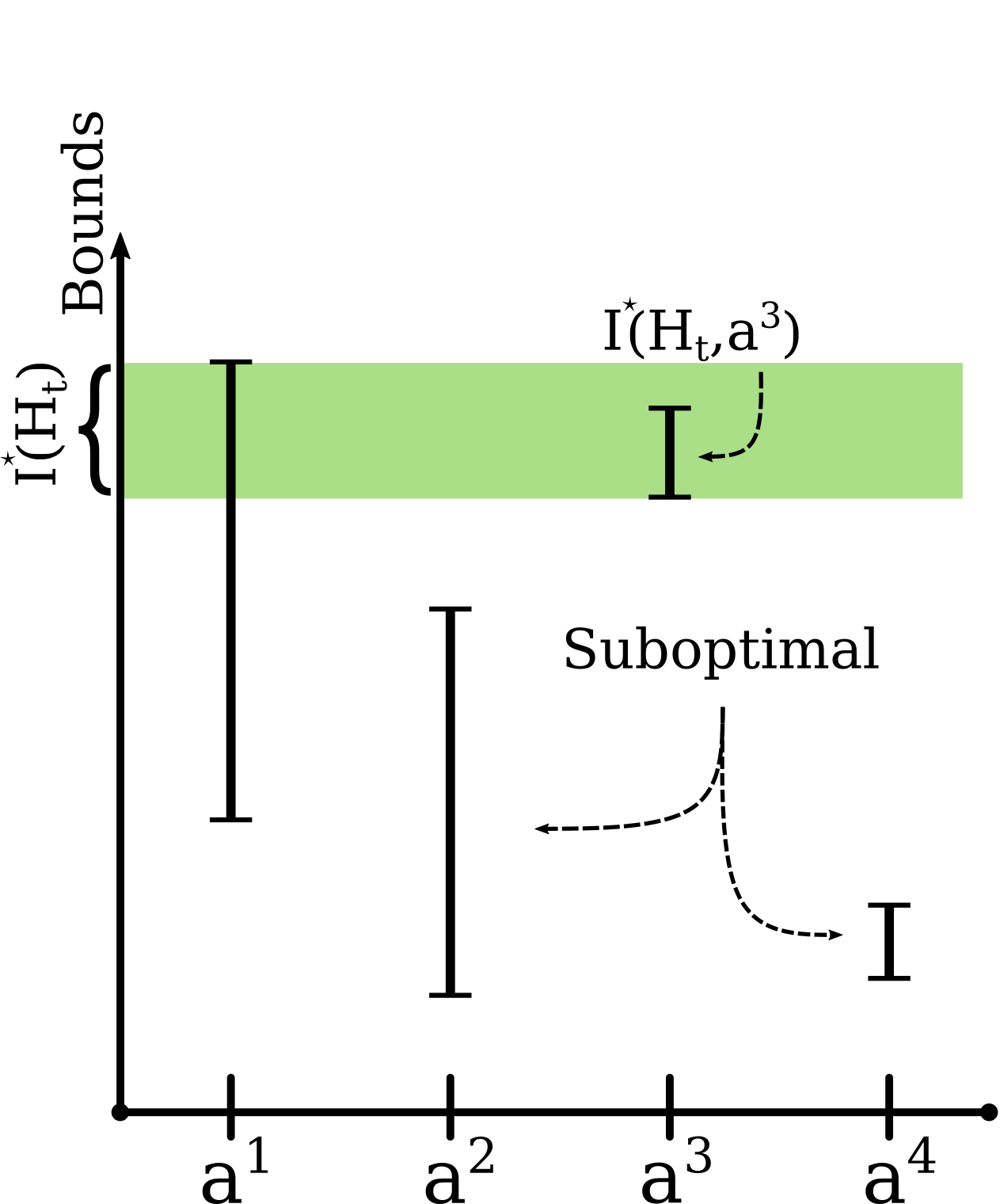}
	\caption{Bound intervals for different actions. The optimal value function
	is guaranteed to be between the maximal lower and upper bounds. As a result,
	actions $a^2$ and $a^4$ are suboptimal and can be pruned safely.
	}
	\label{fig:intervals}
\end{figure}

Then, we define the interval for each action at the root as,
\begin{equation} \label{eq:interval}
    I^{\star}(H_0, a_0)\in\left[\mathcal{L}_0^{\star}(H_t,a_0), \mathcal{U}_0^{\star}(H_0,a_0) \right],
\end{equation}
and use it as a tool for pruning suboptimal actions once an upper bound of an
 action falls below the best lower bounds amongst other actions within that
 node, see figure \ref{fig:intervals} for an illustration.

%\begin{wrapfigure}{r}{0.5\textwidth}
%\end{wrapfigure} 

State-of-the-art algorithms such as POMCP and DESPOT employ
probabilistic and asymptotic reasoning to approximate the optimal policy, and
lack a mechanism to conclusively determine the suboptimality of an action,
leading to infinite exploration of suboptimal actions. In contrast, utilizing
\eqref{eq:interval} guarantees that once an action is identified suboptimal, it
can be safely excluded from further consideration. Since the bounds can be
integrated with arbitrary exploration methods, it provides a novel mechanism for
pruning with contemporary SOTA algorithms.

Importantly, this approach introduces a practical stopping criterion for the
online tree search process. When the exploration results in only one viable
action remaining at the root, it signifies the identification of the optimal
action. Note that this does note necessitate exhaustive exploration of the
entire tree or complete convergence of the bounds.

\subsubsection{Exploration Strategies} \label{subsubsec:exploration}

One can further utilize the root upper bound to determine the exploration of
actions, the simplified state and observation spaces at run time, which
guarantees convergences to the optimal value function in finite time, which is
novel for online tree search POMDPs solvers to the best of our knowledge. We
define the following deterministic exploration strategy,
\begin{align}
    a_t &= \arg \max_{a\in \mathcal{A}}\{\sum_{\tau_{t}\in \mathcal{T}(H_t)} \bar{\mathbb{P}}(\tau_t) r(x_t, a) + \sum_{z_{t+1}\in \bar{\mathcal{Z}}( H_{t},a)}\mathcal{U}_0^{\star}(H_{t+1}) \label{eq:rootBoundExplor} \\ \notag 
    &+ \mathcal{V}_{\max, t} \left[\sum_{\tau_t\in \mathcal{T}(H_t)}\bar{\mathbb{P}}(\tau_t) - \sum_{\tau_{t+1}\in \mathcal{T}(H_{t},a)}\bar{\mathbb{P}}(\tau_{t+1})\right]\}\\
    z_{t+1} &= \arg \max_{o_{t+1} \in \mathcal{Z}(H_{t},a_t)}\{\mathcal{U}_0^{\star}((H_{t}, a_t, o_{t+1})) - \mathcal{L}_0^{\star}((H_{t}, a_t, o_{t+1}))\}\label{eq:obsExplor} \\ 
    x_{t+1} &= \arg \max_{x \in \mathcal{X}(H_{t+1})} \{\bar{\mathbb{P}}^{\star}((\tau_t, a_t, z_{t+1}, x)) - \sum_{\tau_{T}} \bar{\mathbb{P}}^{\star}(\tau_T \mid \tau_t, a_t, z_{t+1}, x)\}, \label{eq:stateExplor}
\end{align}
where the actions are chosen by the highest upper bound, sometimes referred to
as an \textit{"optimism in face of uncertainty"}, which offers a balance between
exploration and exploitation of actions that are possibly optimal or have high
uncertainty in their value. Observations are chosen based on the maximum gap
between the upper and lower bounds, which results in observations with high
uncertainty in their value. Last, we define $\bar{\mathbb{P}}^{\star}(\tau_t)$
as the probability of a trajectory $\tau_t$ under a policy derived from
recursive action selection as per \eqref{eq:rootBoundExplor}. Subsequently, the
selection of states effectively maximizes the difference in probability between
the individual trajectory density and the aggregate probability of all sampled
trajectories that begin with that particular trajectory.

\begin{corollary}\label{lemma:convergenceToOptimal} 
    Performing exploration based on
    \eqref{eq:rootBoundExplor}, \eqref{eq:obsExplor} and \eqref{eq:stateExplor}
    ensures that the algorithm converges to the optimal value function within a
    finite number of planning iterations.
    \begin{proof}
    The proof is provided in \ref{proof:convergenceToOptimal}.
    \end{proof}
\end{corollary}

Importantly, alternative methods for the state-action-observation exploration
are viable and, if given limited planning time, may offer improved performance
in practice. Corollary \ref{lemma:convergenceToOptimal} suggests one way that is
guaranteed to converge in finite time. We leave the investigation of other
approaches for finite-time convergence using the deterministic bounds for future
research.

Moreover, the bounds suggested in this chapter can be integrated with
established algorithms like POMCP or DESPOT (\cite{Silver10nips,Somani13nips}),
an approach which offers several advantages over the existing algorithms. First,
The quality of their solutions with respect to the optimal value can be assessed
and validated. Second, whenever the bounds at the root of the solver do not
overlap, the planning session can be terminated early with a guarantee of
identifying the optimal action.

\subsection[short]{\MB{Impact of POMDP Characteristics on Deterministic Bounds}}
\VI{\noindent To make explicit how POMDP characteristics affect bound tightness,
we recast the upper--lower gap in terms of simplified-space coverage and local
value spread.

    The analytical bounds depend on trajectory-probability terms such as
  $\sum_{\tau}\bar{\mathbb{P}}(\tau)$ and on the auxiliary bounds
  $\mathcal{V}_{\max,t}, \mathcal{V}_{\min,t}$; starting from the recursive
  definitions in \eqref{def:RecursiveBound}, let\; 
  \[
    \Delta
  V(H_t)\;\triangleq\;\mathcal{V}_{\max}(H_t)-\mathcal{V}_{\min}(H_t),
  \]
  \[
  \delta(H_t)\;\triangleq\;
  \sum_{\tau_t\in\mathcal{T}(H_t)}\bar{\mathbb{P}}(\tau_t)\;-\!\!\!
  \sum_{z_{t+1}\in\bar{\mathcal{Z}}(H_t,\pi_t)}\;
        \sum_{\tau_{t+1}\in\mathcal{T}(H_{t+1})}
          \bar{\mathbb{P}}(\tau_{t+1}),
  \]
  then following the derivation in \ref{lower_upper_gap},
  \begin{align}\label{eq:gap}
    \mathcal{U}^\pi_0(H_t)-\mathcal{L}^\pi_0(H_t)\;=\;
    \Delta V(H_t)\,\delta(H_t)
                 +\!\!\!\!\!\sum_{z_{t+1}\in\bar{\mathcal{Z}}(H_t,\pi_t)}\!\!\!\!\!\left[\mathcal{U}^\pi_0(H_{t+1})-\mathcal{L}^\pi_0(H_{t+1})\right].  
  \end{align}

  Equation~\eqref{eq:gap} indicates that the gap accumulates the value function
  spread $\Delta V(H_k)$ only on the time-steps where the trajectory is not part
  of the simplified set. 
  The attributes govern the magnitude of the bounds are twofold:

\begin{enumerate}%[label=(\roman*)]
  \item \textbf{Coverage probability of the simplified spaces.}  
      An important quantity is the probability mass of trajectories not part of
      the simplified sets, $\bar{\mathcal{X}}\!, \bar{\mathcal{Z}}$, rather than
      the cardinality of the omitted sets.  A large POMDP whose belief is highly
      concentrated within the simplified spaces can therefore yield a smaller
      gap than a small, nearly uniform domain.

    \item \textbf{Magnitude of the local value spread, $\Delta V(H_k)$.}  
    Each time a trajectory leaves the simplified spaces, its contribution to the
    gap is scaled by $\Delta
    V(H_k)=\mathcal{V}_{\max}(H_k)-\mathcal{V}_{\min}(H_k)$, the
    policy-independent spread between the upper and lower \emph{cumulative}
    value bounds at node $H_k$. Sharpening these auxiliary bounds, e.g.\ by using
    informed rollouts or MDP-optimal surrogates, reduces $\Delta V(H_k)$ and
    proportionally narrows the overall gap.

\end{enumerate}

\noindent
\emph{Practical implication.}  Maximizing the probability coverage of the
simplified spaces, followed by tightening the auxiliary bounds to reduce
$\Delta V(H_k)$, yields a more favourable gap-complexity trade-off than a
uniform expansion of the simplified spaces.
}

%%%%%%%%%%%%%%%%%%%%%%%%%%%%%
% Algorithms %
%%%%%%%%%%%%%%%%%%%%%%%%%%%%%
\section{Algorithms}
\begin{algorithm}[H]
    \small
    \caption{\textsc{Algorithm}---$\mathcal{A}$:}
    \begin{minipage}[t]{.5\linewidth}
		\label{alg:algA}
        \centering
        \begin{algorithmic}[1]
			\item[] \textbf{function} \textsc{Search}
			\WHILE{time permits}
				\STATE Generate states $x$ from $b_0$. \label{line:samplePrior}
				\STATE $\tau_0 \xleftarrow{} x$
				\STATE $\bar{\mathbb{P}}_0 \xleftarrow{} b(x=\tau_0\mid h_0)$
				\IF{$\tau_0 \notin \tau(h_0)$}
				\STATE $\bar{\mathbb{P}}(h_0) \xleftarrow{}\bar{\mathbb{P}}(h_0)+
				\bar{\mathbb{P}}_0$
				\ENDIF
				\STATE $\textsc{Simulate}(h_0, D, \tau_0, \bar{\mathbb{P}}_0)$.
			\ENDWHILE
			\RETURN 
        \end{algorithmic}
		\vspace{0.1cm}
        \begin{algorithmic}[1]
			\item[] \textbf{function} \textsc{fwdUpdate($ha, haz$, $\tau_d$,
			$\bar{\mathbb{P}}_\tau$, $x'$)}
			\IF{$\tau_{d} \notin \tau(ha)$} \label{line:IF_tau}
			\STATE $\tau(ha) \xleftarrow{} \tau(ha) \cup \{\tau_{d}\}$
			\STATE $\bar{R}(ha) \xleftarrow{} \bar{R}(ha) +
			\bar{\mathbb{P}}_{\tau} \cdot r(x,a)$ \label{line:rewardCalc}
			\ENDIF
			\STATE $\tau_{d} \xleftarrow{} \tau_{d} \cup \{x'\}$
			\STATE $\bar{\mathbb{P}}_{\tau} \xleftarrow{}
			\bar{\mathbb{P}}_{\tau}\cdot Z_{z\mid x'}\cdot T_{x'\mid x, a}$ \label{line:likelihoodProd}
			\IF{$\tau_{d} \notin \tau(haz)$}
			\STATE $\bar{\mathbb{P}}(haz) \xleftarrow{}\bar{\mathbb{P}}(haz) + 
			\bar{\mathbb{P}}_{\tau}$
			\STATE $\tau(haz) \xleftarrow{} \tau(haz) \cup \{\tau_{d}\}$
			\ENDIF
			\RETURN 
        \end{algorithmic}
    \end{minipage}
    \begin{minipage}[t]{.5\linewidth}
		\vspace{0.001cm}
        \centering
        \begin{algorithmic}[1]
			\item[] \textbf{function} $\textsc{Simulate}(h, d, \tau_d, \bar{\mathbb{P}}_d)$
			\IF{$d = 0$}
			\RETURN \label{line:endSim}
			\ENDIF
			\STATE Select action $a$. \label{line:Aexploration}
			\STATE Generate next states and observations,
			$x',z$.\label{line:gen} \label{line:SOexploration}
			\STATE $\tau_d, \ \bar{\mathbb{P}}_\tau
			\xleftarrow{}$\textsc{fwdUpdate($ha, haz, \tau_d, \bar{\mathbb{P}}_\tau, x'$)}
			\STATE Select next observation $z$.
			\STATE \textsc{Simulate}$(haz, d-1, \tau_d, \bar{\mathbb{P}}_{\tau})$
			\STATE \textsc{bwdUpdate($h, ha, d$)}
			\RETURN 
        \end{algorithmic}
		\vspace{0.1cm}
        \begin{algorithmic}[1]
			\item[] \textbf{function} \textsc{bwdUpdate($h, ha, d$)}
			\STATE $\epsilon(ha) =
			\gamma^{D-d}V_{\max, d}(\bar{\mathbb{P}}(h) - \bar{\mathbb{P}}(ha))
			+\gamma^{D-d-1} \cdot V_{\max, d+1}(\bar{\mathbb{P}}(ha) -
			\sum\limits_{z\mid ha} \bar{\mathbb{P}}(haz))$ \label{line:bounds}
			\STATE $U(ha)\! =\! \bar{R}(ha) + \gamma\! \sum_{z\mid ha}\!
			U(haz) + \epsilon(ha)$ \label{line:calcBounds}
			\STATE $L(ha)\! =\! \bar{R}(ha) + \gamma\! \sum_{z\mid ha}\!
			L(haz) - \epsilon(ha)$ 
			\STATE $U(h) \xleftarrow{} \max_{a'} \{U(ha')\}$
			\STATE $L(h) \xleftarrow{} \max_{a'} \{L(ha')\}$
			\RETURN 
        \end{algorithmic}
    \end{minipage}
\end{algorithm}

In this section we aim to describe how to fit our bounds to a blueprint of a
general algorithm, named \textsc{Algorithm}---$\mathcal{A}$, which serves as an
abstraction to many existing algorithms. Then, we explicitly describe two
algorithms, DB-POMCP, an adaptation to POMCP that uses UCB for
exploration, and our deterministic bounds for decision-making, and RB-POMCP, a particle-based solver that utilizes the bounds both for decision-making and exploration.

To compute the deterministic bounds, we utilize Bellman's update and optimality
criteria. This approach naturally fits dynamic programming approaches such as
DESPOT \citep{Ye17jair} and AdaOPS \citep{Wu21nips}. However, it may also be
attached with algorithms that rely on Monte-Carlo estimation, such as POMCP
\citep{Silver10nips}, by viewing the search tree as a policy tree.

While the analysis presented in section \ref{sec:analysis} is general and
independent of the selection mechanism of the states or observations, we focus
on sampling as a way to choose the simplified states at each belief node and the
observations to branch from. Furthermore, the selection of the subspaces
$\bar{\mathcal{X}}, \bar{\mathcal{Z}}$ need not be fixed, and may change over
the course of time, similar to state-of-the-art algorithms, such as
\cite{Hoerger21icra,Silver10nips,Somani13nips,Sunberg18icaps,Wu21nips}.
Alternative selection methods may also be feasible, as sampling from the correct
distribution is not required for the bounds to hold. Importantly, attaching our
bounds to arbitrary exploration mechanism certifies the algorithm solution with
deterministic bounds to the optimal solution, and may result in an improved
decision making, as will be shown in the experimental section.

\textsc{Algorithm}---$\mathcal{A}$ is outlined in algorithm \ref{alg:algA}. For
clarity of exposition, we assume the following; at each iteration a single state
particle is propagated from the root node to the leaf (line
\ref{line:samplePrior} of function \textsc{Search}). The selection of the next
state and observations are done by sampling from the observation and transition
models (line \ref{line:gen}), and each iteration ends with the full horizon of
the POMDP (lines \ref{line:endSim}). However, none of these are a restriction of
our approach and may be replaced with arbitrary number of particles, arbitrary
state and observation selection mechanism and a single or multiple expansions of
new belief nodes at each iteration.

To compute the bounds, we require both the state trajectory, denoted as
$\tau$, and its probability value, $\mathbb{P}_{\tau}$. We use the state
trajectory as a mechanism to avoid duplicate summation of an already accounted
for probability value and is utilized to ascertain its uniqueness at a belief
node. The probability value, $\mathbb{P}_{\tau}$, is the likelihood of visiting
a trajectory $\tau=\{x_0, a_0, x_1, z_1, \ldots, a_{t-1}, x_t, z_t\}$ and is
calculated as the product of the prior, transition and observation likelihoods
(line \ref{line:likelihoodProd}). If a trajectory was not previously observed in
a belief node, its reward value is multiplied by the likelihood of the
trajectory. Each trajectory likelihood is maintained as part of a cumulative sum
of all visited trajectories in the node. This cumulative sum is then used to
calculate the upper and lower bounds, which are shown in lines
\ref{line:bounds}-\ref{line:calcBounds}. The term computed in line
\ref{line:bounds} represents the loss of holding only a subset of the states in
node $ha$ from the set in node $h$, plus the loss of having only a partial set
of posterior nodes and a subset of their states. $V_{\max,d}$ represents an
upper bound for the value function. A simple bound on the value function can be
$V_{\max,d}=\mathcal{R}_{\max} \cdot (D-d)$, but other more sophisticated bounds
may also be used. In the experimental section we show that despite the
additional overhead, utilizing the deterministic bounds, \eqref{eq:lowerBound} and \eqref{eq:upperBound}, within the actual decision-making
improves the results of the respective algorithms.

\subsection[short]{DB-POMCP} 
DB-POMCP uses theorem \ref{thm:valueFuncSimplifiedBoth} for decision-making once an optimal action was found or at time-out given limited planning time. 
 In aligning Algorithm \ref{alg:algA} with the POMCP
framework, the action exploration process determined by the
Upper Confidence Bounds for Trees (UCT) criterion,
\begin{equation}
	UCT(H_t, a_t) = \hat{Q}^{mean}(H_t, a_t) + c\sqrt{\frac{log(N(H_t))}{N(H_t, a_t)}},
\end{equation}
where $\hat{Q}^{mean}$ is the average of the cumulative sums obtained from
sampled explorations, and $c$ is a tunable constant that trades-off exploration and exploitation during planning. Following this criterion, each state and observation is
then sampled according to their respective transition and observation models.
The original POMCP method, as discussed in \cite{Silver10nips}, employs
Monte-Carlo rollouts for value estimation and refrains from adding new nodes
during these rollouts. During our evaluations we saw a negligible difference in
performance, thus we avoid presenting rollouts to algorithm \ref{alg:algA} for
simplicity. However, DB-POMCP supports both settings.

\subsection[short]{RB-POMCP}  \label{subsec:rbpomcp}

Root-Bounded POMCP (RB-POMCP) differs from DB-POMCP in that it uses a different exploration method. We denote it
RB-POMCP to emphasize that the bounds hold only in the root node, and are not
valid for any node along the tree, yet unlike DB-POMCP the bounds are used for exploration in
any part of the tree. The RB-POMCP methodology draws inspiration from the
Monte-Carlo approach suggested the original POMCP algorithm and
innovates by incorporating upper and lower bounds, as defined in
\eqref{eq:optimalRecursive} and \eqref{eq:optimalRecursive2}, to guide both the exploration and the decision-making
processes.

The RB-POMCP framework is constructed based on the structure outlined in Algorithm
\ref{alg:algA}, which necessitates specific implementations for abstract state,
action, and observation exploration functions. In our approach, we opt for an
approximation to the exploration mechanism proposed in 
section \ref{subsubsec:exploration}. More precisely, while we adhere to the action
exploration strategy described in the lemma, we simplify the observation and
state exploration components by employing basic Monte-Carlo sampling techniques,
akin to those used in the standard POMCP algorithm. This modification is
intended to enhance the algorithm's planning efficiency without compromising the
integrity of the algorithm bounds. The remainder of the RB-POMCP algorithm
adheres closely to the procedures specified in Algorithm \ref{alg:algA}.
Additionally, we use pruning and stopping criteria, as described in
\ref{subsec:pruningStoppingCriteria}.

\subsection[short]{Time complexity} \label{subsec:timeComplexity}
The cost of updating a posterior node depends on the underlying solver. For
dynamic programming methods, such as DESPOT and AdaOPS, the extra bookkeeping
required by our bounds is negligible, so the original time complexity is
essentially unchanged. For Monte Carlo methods, such as POMCP, the
baseline complexity is $O(|\mathcal{A}|)$ related mainly to the
action-selection; A naive implementation of our bounds adds another linear
complexity term, making it $O(|\mathcal{A}| + |\bar{\mathcal{Z}}|)$ due to the
summation over the simplified observation space described in Corollary
\ref{lemma:udbOptimality} and algorithm \ref{alg:algA}, line \ref{line:bounds}
of \textsc{bwdUpdate} function. 

However, we show that our added bounds can still be updated in $O(|\mathcal{A}|)$
by storing two additional scalar bookkeeping values for each node. Namely,
incremental visitation probability, $\Delta \mathbb{P}^{(i)}(\cdot)$, and the
change to child's upper bound, $\Delta
\mathcal{U}^{(i)}(\cdot)$. 

Each POMCP visit touches a single observation child, so only that branch must be
updated. The bounds are therefore updated incrementally,
\begin{gather} \label{eq:incrementalUpdate}
	\mathcal{U}_{0}^{( i)}( H_{t} ,a_{t}) =\mathcal{U}_{0}^{( i-1)}( H_{t} ,a_{t}) +\Delta \overline{\mathbb{P}}^{( i)}\left( \tau _{t}^{( i)}\right)\left[ r\left( x_{t}^{( i)} ,a_{t}\right) +V_{max,t}\right] \\
	+\Delta \mathcal{U}_{0}^{( i)}\left( H_{t+1}^{( i)}\right) -\Delta \overline{\mathbb{P}}^{( i)}\left( \tau _{t+1}^{( i)}\right) V_{max,t} \notag
\end{gather}
and,
\begin{equation}
	\mathcal{U}_{0}^{( i)}( H_{T}) =\underset{a_{t} \in \mathcal{A}}{\max}\left\{\mathcal{U}_{0}^{( i)}( H_{t} ,a_{t})\right\}.
\end{equation}

Because only one branch is modified, the per-visit overhead
remains $O(|\mathcal{A}|)$. The full algebraic derivation is provided in
\Cref{lower_upper_gap}.

During each visitation to a node, the trajectory bookkeeping $\tau_d \in
\tau(ha)$ shown in algorithm \ref{alg:algA} line \ref{line:IF_tau}, is used to
determine whether a specific trajectory has already been encountered at the
current node. This verification process can potentially result in an added
linear complexity of $O(D)$, where $D$ represents the planning horizon. However,
this overhead can be circumvented by assigning a unique ID value to each
trajectory at the previous step and subsequently checking whether a pair,
comprising the ID value and the new state, has already been visited. This
approach reduces the overhead to an average time complexity of $O(1)$ by
utilizing hash maps efficiently.

\MB{While the asymptotic time complexity remains similar, the deterministic
certificates add a constant-factor bookkeeping cost per node, due to hash
lookups and updates of bounds ($O(|A|)$ worst-case plus $O(1)$ on average for
algorithm \ref{alg:algA}). Practical wall-clock effects depend on implementation
details and engineering optimizations, so we confine this section to the
algorithmic analysis; systematic profiling is orthogonal to our contribution and
left to future work.}

\section{Experiments}\label{sec:experiments}
Our primary contribution is of a theoretical nature, yet we conducted
experiments to evaluate the practical applicability of our proposed
methodologies. Initially, we adopted a hybrid strategy, such as DB-POMCP, by incorporating our
deterministic bounds exclusively for the decision-making, while relying on existing exploration
strategies such as POMCP and DESPOT. Essentially, this approach enhances the
POMCP and DESPOT frameworks by equipping them with mechanisms that ensure
bounded sub-optimality. In a subsequent experimental setup, we applied the
deterministic bounds to both the exploration and decision-making phases, based
on the methodologies outlined in section \ref{subsec:rbpomcp}. We then compared
the empirical performance of using the deterministic bounds solely for
decision-making to the baseline algorithms without the
incorporation of any deterministic bounds. Our findings indicate that while the
application of deterministic bounds to decision-making can enhance performance,
this strategy becomes less effective in identifying the optimal action as the
complexity of the POMDP increases. Conversely, when the deterministic bounds are
applied to both exploration and decision-making (section \ref{subsec:rbpomcp}), the results demonstrate a
linear increase in planning time proportional to the size of the POMDP,
indicating better scalability.

\subsection[short]{Deterministic-Bounds for Decision-Making}
In this subsection, we focus on the application of deterministic bounds
exclusively for decision-making. This approach involves using a predefined
exploration strategy during the planning phase, but making the final action
selection based on the deterministic bounds as shown in
\eqref{eq:optimalRecursive}. The comparative results for the standard and
deterministically-bounded versions of the POMCP and DESPOT algorithms are
presented in Table \ref{tab:certify}. These versions, labeled DB-POMCP and
DB-DESPOT, adhere to the original exploration criteria of their respective
algorithms but select actions based on the highest lower bound, as specified in
\eqref{eq:lowerBound}.

Our experimental analysis reveals that, in addition to offering a level of
optimality certification for the chosen actions, utilizing deterministic bounds
for action selection can enhance the expected cumulative reward. It is important
to note, however, that this method does not always lead to better outcomes.
Specifically, it may not be advantageous in situations where the highest lower
bound is less than other available upper bounds (for instance, comparing actions
$a^1$ and $a^3$ in figure \ref{fig:intervals}). \VI{In practice, when the bounds
overlap, using them for action selection provides only heuristic guidance and
may be incorrect. In our experiments, we used uniform bounds accross all
future belief states, which are not useful for guiding the policy towards better
actions. In fact, the more symmetric the bounds are at different nodes, the more
belief nodes the algorithm will have to visit to distinguish between the
attractiveness of different actions. While this can be improved by considering
informed bounds, e.g. MDP-optimal for upper bounds or rollout-based lower
bounds, we leave it for future work. When using uninformed bounds, a large POMDP
with large state, observation or action spaces, leads to more loose bounds, which
in turn reduces the effectiveness of our approach.} This limitation is evident
in the results for the Laser Tag POMDP, a considerably larger problem compared
to the other POMDPs evaluated, where the deterministic bounds did not yield
performance improvements.

\begin{table}[htbp]
  \centering
  \caption{Performance comparison with and without deterministic bounds, for
  short horizon, $H=5$.}
  \label{tab:certify}
  \resizebox*{\textwidth}{!}{%
  \begin{tabular}{l*{4}{S[
    table-format=-2.2,
    separate-uncertainty = true
    % uncertainty-mode = separate % <-- siunitx v3 equivalent
  ]}}
    \toprule
    & \multicolumn{1}{c}{Tiger POMDP}
    & \multicolumn{1}{c}{Laser Tag}
    & \multicolumn{1}{c}{Discrete Light Dark}
    & \multicolumn{1}{c}{Baby POMDP} \\
    \textit{Cardinalities $(|S|,|A|,|O|)$} &
    \multicolumn{1}{c}{(2, 3, 2)} &
    \multicolumn{1}{c}{(5930, 5, $\sim\!1.5\!\times\!10^6$)} &
    \multicolumn{1}{c}{(122, 5, 21)} &
    \multicolumn{1}{c}{(2, 2, 2)} \\
    \midrule
    DB-DESPOT (ours) & 3.74 \pm 0.48 & -5.29 \pm 0.14 & -5.29 \pm 0.01 & -3.92 \pm 0.56 \\
    AR-DESPOT & 2.82 \pm 0.55 & -5.10 \pm 0.14 & -61.53 \pm 5.80 & -5.40 \pm 0.85 \\
    \midrule
    DB-POMCP (ours) & 3.01 \pm 0.21 & -3.97 \pm 0.24 & -3.70 \pm 0.82 & -4.48 \pm 0.57 \\
    POMCP & 2.18 \pm 0.76 & -3.92 \pm 0.27 & -4.51 \pm 1.15 & -5.39 \pm 0.63 \\
    \bottomrule
  \end{tabular}%
  }
\end{table}

\subsection[short]{Root-Bounds for Decision-Making and Exploration}

\begin{table}[htbp]
    \centering
    \caption{Performance comparison with and without deterministic bounds, for medium horizon, $H=15$.}
    \label{tab:pomcpVersions}
    \resizebox*{\textwidth}{!}{%
    \begin{tabular}{l*{4}{S[table-format=-1.2,
                                table-space-text-post=\,(0)]}}
      \toprule
      Algorithm & {Tiger POMDP} & {Rock Sample} & {Navigate to Goal} & {Baby POMDP}\\
      \textit{Cardinalities $(|S|,|A|,|O|)$} &
      \multicolumn{1}{c}{(2, 3, 2)} &
      \multicolumn{1}{c}{(1801, 8, 3)} &
      \multicolumn{1}{c}{(25, 5, 25)} &
      \multicolumn{1}{c}{(2, 2, 2)} \\
      \midrule
      RB-POMCP (ours) & 1.53 \pm 0.76 & 8.50\pm 0.22 & 61.21 \pm 0.71 & -11.97 \pm 0.27 \\
      % \midrule
      % DB-DESPOT (ours) & 0.00 \pm 0.00 & 0.00 \pm 0.00 & 0.00 \pm 0.00 & 0.00 \pm 0.00 \\
      % AR-DESPOT & 8.33 \pm 0.68 & 8.26 \pm 0.22 & 51.59 \pm 0.68 & -12.88 \pm 0.27 \\
      % \midrule
      DB-POMCP (ours) & -1.05 \pm 0.15 & 7.86 \pm 0.21 & 62.37 \pm 0.75 & -12.13 \pm 0.22 \\
      POMCP & -5.59 \pm 0.24 & 5.69 \pm 0.20 & 68.45 \pm 0.69 & -12.49 \pm 0.27 \\
      \bottomrule
    \end{tabular}%
    }
\end{table}

The performance outcomes presented in Table \ref{tab:pomcpVersions} reveal that
the RB-POMCP algorithm typically matches or surpasses the standard POMCP in
various tested environments, except for the Navigate to Goal POMDP scenario. The
limited performance in this particular context can be attributed to the nature
of RB-POMCP's exploration strategy, which is designed to assure optimality over
extended planning periods but does not inherently guarantee enhanced results
within limited planning durations. Unlike probabilistic algorithms that leverage
statistical concentration inequalities-such as the Hoeffding inequality employed
in the Upper Confidence Bounds for Trees (UCT) \cite{Kocsis06ecml} exploration
mechanism of POMCP-RB-POMCP adopts a more cautious strategy. This approach
entails considering both worst-case and best-case scenarios to establish a
deterministic link with the optimal value may not always translate to superior
immediate performance due to its conservative nature. 

% \MB{However, these can be
% improved by considering a more informed bounds, instead of the uniform bounds
% applied here. We also note that Navigate to Goal POMDP is a highly symmetric
% problem, where different actions may lead to similar value function. This, in
% turn, forces the algorithm to visit more nodes more often to distinguish between
% bounds, which may hurt performance.} \MB{Another factor which may affect the
% perfomance of the algorithm are the uniformity of the various distributions, which
% imply a non-negligible likelihood of a specific state or observation, which may
% force the algorithm have to visit before pruning the branch.}

\subsection[short]{Planning for optimal action}
\begin{figure}[h]
	\centering
	\includegraphics[width=0.6\linewidth]{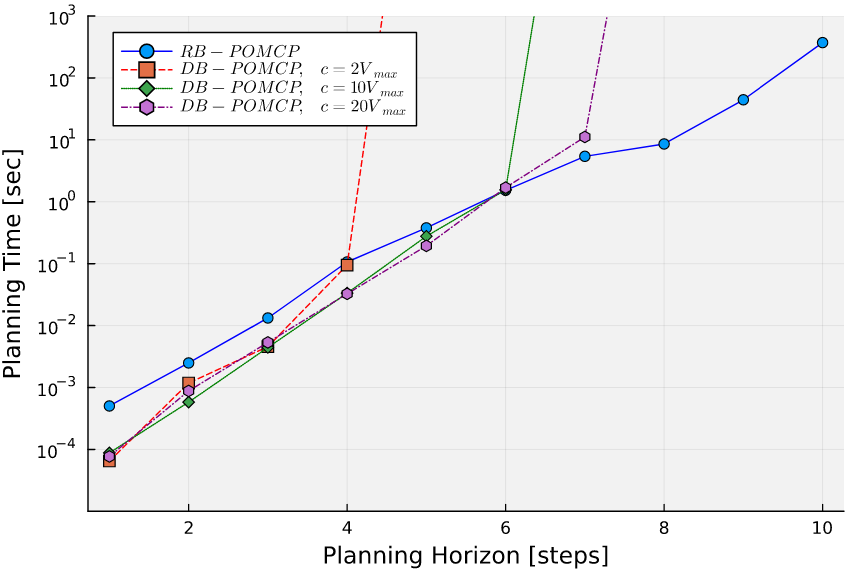}
	\caption{The graphs show the measured planning time for RB-POMCP and DB-POMCP
	to find the optimal action for Rock Sample under different UCT
	coefficient values. Guaranteeing the optimal action made possible by using the
	bounds in corollary \ref{lemma:udbOptimality}.  All simulation runs were capped at
	3,600 seconds.}
    \label{fig:time_plot}
\end{figure}

To highlight the differences between RB-POMCP and DB-POMCP, we examined each algorithm's planning time to deterministically identify the optimal value, as depicted in Figure \ref{fig:time_plot}. Notably, conventional state-of-the-art algorithms, such as POMCP and DESPOT, cannot deterministically identify the optimal action within a finite timeframe and are thus not considered in this analysis. 

DB-POMCP incorporates the Upper Confidence Bounds for Trees (UCT) method for exploration. However, its exploration strategy lacks awareness of the deterministic bounds of the optimal value function, leading to insufficient guidance toward actions that may be optimal. Despite significantly increasing the exploration coefficient beyond the values suggested in previous works \cite{Silver10nips, Sunberg18icaps}, our findings, as presented in Figure \ref{fig:time_plot}, demonstrate that the exploration bonus diminishes too rapidly, effectively limiting further exploration of potentially optimal actions. While UCT, in theory, explores the belief tree indefinitely, in practical scenarios, the exploration rate of new branches diminishes exponentially over time, making it less effective in environments where identifying the optimal action in a reasonable time is crucial. Conversely, RB-POMCP directly utilizes upper and lower bounds information, facilitating a more targeted search for the optimal value. This approach leads to a planning duration that scales linearly with the problem size, as evidenced in Figure \ref{fig:time_plot}, highlighting its efficiency in identifying optimal actions within a finite timeframe.

\subsection[short]{Technical Details}
The implementation of our algorithm written in the Julia programming language,
using the Julia POMDPs package for evaluation and the vanilla POMDP versions,
provided by \cite{egorov2017pomdps}. This package primarily supports infinite
horizon problems; however, we modified it to also handle finite-horizon POMDPs.
The experiments were conducted on a computing platform consisting of an Intel(R)
Core(TM) i7-7700 processor with 8 CPUs operating at 3.60GHz and 15.6 GHz. The
hyper-parameters for the POMCP and AR-DESPOT solvers, and further details about
the POMDPs used for our experiments are detailed in the appendix.

\section{Discussion and future work}
\VI{The principal advantage of the proposed algorithm is its explicit and
deterministic upper and lower bounds on the value function, quantifying the
maximum deviation of the current policy from optimality. These bounds support a
principled early termination by choosing a user-specified tolerance~$\epsilon$,
guaranteeing an $\epsilon$-optimal policy and pruning of provably irrelevant
action branches - capabilities that, to our knowledge, existing online POMDP
planners lack. Moreover, they allow one to utilize the bounds for deterministic
limits on the probability of catastrophic outcomes, an essential feature for
safety-critical tasks. 

However, the present implementation has few shortcomings. First, while not
limited by the theoretical derivations, our implementation considers naive
bounds on the value function, that is, best-or-worst possible rewards at each
unvisited history or state. This results in relatively loose bounds which
hinders scalability to large POMDPs. Tighter alternatives, such as
history-dependent relaxations or bounds derived from the MDP-optimal value
function commonly used in other POMDP solvers would alleviate this issue. We
believe that this opens a new avenue of research, that focuses on general,
efficient value function bounds which are easily applicable to the present
deterministic-bounds algorithm (see $V_{max,d}$ in
\textsc{Algorithm}-$\mathcal{A}$ line \ref{line:bounds}). Second, it adds
implementation complexity, especially apparent compared simple algorithms like
POMCP. Last, the bounds add bookkeeping overhead in both time and memory when
attached to existing algorithms, although section \ref{subsec:timeComplexity}
shows that the asymptotic complexity can be reduced to match existing
algorithms, and pruning suboptimal actions may further reduce the time
efficiency.}

%%%%%%%%%%%%%%%%%%%%%%%%%%%%%
% Conclusions %
%%%%%%%%%%%%%%%%%%%%%%%%%%%%%
\section{Conclusions}
This work addresses the computational challenges of decision-making under
uncertainty, typically formalized as Partially Observable Markov Decision
Processes (POMDPs). Our objective is to bridge the theoretical gap between the
quality of solutions obtained from approximate solvers and the generally
intractable optimal solutions. We present a novel methodology that guarantees
anytime, deterministic bounds for approximate POMDP solvers. We  achieve this by
defining a simplified POMDP, that utilizes only a subset of the state and
observation spaces to alleviate the computational burden. We establish a
theoretical relationship between the optimal value function, which is
computationally intensive, and a more tractable value function obtained using
the simplified POMDP. Based on the theoretical derivation, we suggest the use of
the deterministic bounds to govern the exploration, while being theoretically
guaranteed to converge to the optimal value in finite time. Building upon this
theoretical framework, we show how to integrate the bounds with a general
structure of common state-of-the-art algorithms. Additionally, we leverage our
deterministic bounds to develop an early stopping criterion that identifies
convergence to the optimal value, a novel capability that is not possible with
existing probabilistic bounds. We introduce two algorithms that incorporate the
suggested bounds, named DB-POMCP and RB-POMCP. DB-POMCP exploits the
deterministic relationship for decision-making, while RB-POMCP uses the bounds
for both decision-making and exploration. Finally, we evaluate the practical use
of our approach by comparing the suggested algorithms to state-of-the-art
algorithms, demonstrating their effectiveness.

\section*{Acknowledgments}
This work was supported by the Israel Science Foundation (ISF) and by US NSF/US-Israel BSF.

\bibliographystyle{plainnat}
\bibliography{refs}
%%%%%%%%%%%%%%%%%%%%%%%%%%%%%%%%%%%%%%%%%%%%%%%%%%%%%%%%%%%%

%%%%%%%%%%%%%%%%%%%%%%%%%%%%%
% Appendix %
%%%%%%%%%%%%%%%%%%%%%%%%%%%%%
\appendix
\section{Appendices}

% \tableofcontents
\subsection{Mathematical Analysis} 

We start by restating the definition of the simplified value function,
\begin{align} %\label{def:simplifiedValueFunc}
    \bar{V}^\pi(\bar{b}_t)&\triangleq r(\bar{b}_t, \pi_t) + \bar{\mathbb{E}}\left[\bar{V}(b_t)\right]\\
    &=\sum_{x_t}\bar{b}(x_t)r(x_t,\pi_t) + \sum_{z_t}\bar{\mathbb{P}}(z_{t+1}\mid
    H^-_{t+1})\bar{V}(\bar{b}(z_{t+1})),
\end{align}
\label{app:proof-thm-valueFuncSimplifiedObs}
\subsubsection{{\Cref{thm:valueFuncSimplifiedObs}}}
    Let $b_t$ belief state at time $t$, and $T$ be the last time step of the POMDP.
    Let $V^{\pi}(b_t)$ be the theoretical value function by following a policy
    $\pi$, and let $\bar{V}^{\pi}(b_t)$ be the simplified value function, 
    as defined in \eqref{def:simplifiedValueFunc}, by following the same policy.
    Then, for any policy $\pi$, the difference between the theoretical and
    simplified value functions is bounded as follows,
    \begin{equation} %\label{eq:thm}
        \left|V^\pi(b_t) \! - \! \bar{V}^\pi(b_t)\right|\leq \! \mathcal{R}_{\max} \! \! \! \sum _{\tau=t+1}^{T} \!\! \left[ 1\! -\! \! \! \sum_{z_{t+1:{\tau}}}\sum
        _{x_{t:{\tau}}}b( x_{t}) \! \! \prod _{k=t+1}^{\tau}\bar{\mathbb{P}}( z_{k} \mid x_{k})
        \mathbb{P}( x_{k}\mid x_{k-1} ,\pi _{k-1})\right]\triangleq \epsilon_z^\pi(b_t).
    \end{equation}
    \begin{proof}%[Proof of \Cref{thm:valueFuncSimplifiedObs}]
        For notational convenience, we derive the bounds for the value function
        by denoting the prior belief as $b_0$,
        \begin{equation}
            V_0^\pi(b_0)=\mathbb{E}_{z_{1:T}}\left[\sum_{t=0}^T r(b_t,a_t)\right]
        \end{equation}
        applying the belief update equation,
        \begin{align}
        V_0^\pi(b_0) &=\sum _{z_{1:T}}\prod _{\tau =1}^{T}\mathbb{P}\left( z_{\tau } \mid H_{\tau }^{-}\right)\sum _{t=0}^{T}\left[\sum _{x_{t}}\frac{\mathbb{P}( z_{t} \mid x_{t})\sum _{x_{t-1}}\mathbb{P}( x_{t} \mid x_{t-1} ,\pi _{t-1}) b_{t-1}}{\mathbb{P}\left( z_{t} \mid H_{t}^{-}\right)} r( x_{t} ,a_{t})\right]\\
        &=\sum _{z_{1:T}}\prod _{\tau =1}^{T}\mathbb{P}\left( z_{\tau } \mid H_{\tau }^{-}\right)\sum _{t=0}^{T}\left[\sum _{x_{0:t}}\frac{\ \prod _{k=1}^{t} \ \mathbb{P}( z_{k} \mid x_{k})\mathbb{P}( x_{k} \mid x_{k-1} ,\pi _{k-1}) b( x_{0})}{\prod _{\tau =1}^{t}\mathbb{P}\left( z_{\tau } \mid H_{\tau }^{-}\right)} r( x_{t} ,a_{t})\right]\\
        &=\sum _{t=0}^{T}\sum _{z_{1:T}}\sum _{x_{0:T}}\prod _{k=1}^{t} \ \mathbb{P}( z_{k} \mid x_{k})\mathbb{P}( x_{k} \mid x_{k-1} ,\pi _{k-1}) b( x_{0}) r( x_{t} ,a_{t})
        \end{align}
        which applies similarly to the simplified value function, 
        \begin{equation}
            \bar{V}_0^\pi(b_0)=\sum _{t=0}^{T}\sum _{z_{1:T}}\sum _{x_{0:T}}\prod _{k=1}^{t} \ \bar{\mathbb{P}}( z_{k} \mid x_{k})\mathbb{P}( x_{k} \mid x_{k-1} ,\pi _{k-1}) b( x_{0}) r( x_{t} ,a_{t}).
        \end{equation}
        We begin the derivation by focusing on a single time step, $t$, and
        later generalize to the complete value function.
        \begin{align}
&| \mathbb{E}_{z_{1:t}}[ r( b_{t})] -\bar{\mathbb{E}}_{z_{1:t}}[ r(\bar{b}_{t})]| \\
=&| \sum _{z_{1:t}}\sum _{x_{0:t}}[\prod _{k=1}^{t} \ \mathbb{P}( z_{k} \mid x_{k})\mathbb{P}( x_{k} \mid x_{k-1} ,\pi _{k-1}) b( x_{0}) r( x_{t}) \\ \notag
&-\prod _{k'=1}^{t} \ \bar{\mathbb{P}}( z_{k'} \mid x_{k'})\mathbb{P}( x_{k'} \mid x_{k'-1} ,\pi _{k'-1}) b( x_{0}) r( x_{t})]| \\
\leq& \sum _{z_{1:t}}\sum _{x_{0:t}}\left| r( x_{t})\left[\prod _{k=1}^{t}\mathbb{P}( z_{k} \mid x_{k})\mathbb{P}( x_{k} \mid x_{k-1} ,\pi _{k-1}) b( x_{0}) -\prod _{k'=1}^{t} b( x_{0}) \ \bar{\mathbb{P}}( z_{k'} \mid x_{k'})\mathbb{P}( x_{k'} \mid x_{k'-1} ,\pi _{k'-1})\right]\right| \\
=&\sum _{z_{1:t}}\sum _{x_{0:t}}| r( x_{t})| \left[\prod _{k=1}^{t}\mathbb{P}( z_{k} \mid x_{k})\mathbb{P}( x_{k} \mid x_{k-1} ,\pi _{k-1}) b( x_{0}) -\prod _{k'=1}^{t} b( x_{0}) \ \bar{\mathbb{P}}( z_{k'} \mid x_{k'}) \ \mathbb{P}( x_{k'} \mid x_{k'-1} ,\pi _{k'-1})\right]
        \end{align}
        where the second transition is due to triangle inequality, the third
        transition is equality by the construction, i.e. using the simplified
        observation models imply that the difference is nonnegative. We add
        and subtract, followed by rearranging terms,
        \begin{align}
            =&\sum _{z_{1:t}}\sum _{x_{0:t}}| r( x_{t})| \\
&[\prod _{k=1}^{t}\mathbb{P}( z_{k} ,x_{k} \mid x_{k-1} ,\pi _{k-1}) b( x_{0}) -\prod _{k=1}^{t-1} b( x_{0})\bar{\mathbb{P}}( z_{k} ,x_{k} \mid x_{k-1} ,\pi _{k-1})\mathbb{P}( z_{t} ,x_{t} \mid x_{t-1} ,\pi _{t-1})\notag\\
&+\prod _{k=1}^{t-1}b( x_{0})\bar{\mathbb{P}}( z_{k} ,x_{k} \mid x_{k-1} ,\pi _{k-1})\mathbb{P}( z_{t} ,x_{t} \mid x_{t-1} ,\pi _{t-1}) -\prod _{k'=1}^{t} b( x_{0})\bar{\mathbb{P}}( z_{k'} ,x_{k'} \mid x_{k'-1} ,\pi _{k'-1})]\notag\\
=&\sum _{z_{1:t}}\sum _{x_{0:t}}| r( x_{t})| \Bigl\{\\
&\mathbb{P}( z_{t} ,x_{t} \mid x_{t-1} ,\pi _{t-1})\left[\prod _{k=1}^{t-1}\mathbb{P}( z_{k} ,x_{k} \mid x_{k-1} ,\pi _{k-1}) b( x_{0}) -\prod _{k=1}^{t-1} b( x_{0})\bar{\mathbb{P}}( z_{k} ,x_{k} \mid x_{k-1} ,\pi _{k-1})\right] \notag\\
&+\prod _{k=1}^{t-1}b( x_{0})\bar{\mathbb{P}}( z_{k} ,x_{k} \mid x_{k-1} ,\pi _{k-1})[\mathbb{P}( z_{t} ,x_{t} \mid x_{t-1} ,\pi _{t-1}) -\ \bar{\mathbb{P}}( z_{t} ,x_{t} \mid x_{t-1} ,\pi _{t-1})]\Bigr\} \notag
        \end{align}
        applying Holder's inequality,
        \begin{align}
            \leq& \mathcal{R}_{\max}\sum _{z_{1:t}}\sum _{x_{0:t}}\mathbb{P}( z_{t} ,x_{t} \mid x_{t-1} ,\pi _{t-1})\left[ b( x_{0})\prod _{k=1}^{t-1}\mathbb{P}( z_{k} ,x_{k} \mid x_{k-1} ,\pi _{k-1}) -b( x_{0})\prod _{k=1}^{t-1}\bar{\mathbb{P}}( z_{k} ,x_{k} \mid x_{k-1} ,\pi _{k-1})\right]\\
            &+\mathcal{R}_{\max}\sum _{z_{1:t}}\sum _{x_{0:t}}\prod _{k=1}^{t-1}\bar{\mathbb{P}}( z_{k} ,x_{k} \mid x_{k-1} ,\pi _{k-1}) b( x_{0})[\mathbb{P}( z_{t} ,x_{t} \mid x_{t-1} ,\pi _{t-1}) -\ \bar{\mathbb{P}}( z_{t} ,x_{t} \mid x_{t-1} ,\pi _{t-1})]\notag \\
            =&\mathcal{R}_{\max}\sum _{z_{1:t}}\sum _{x_{0:t}}\mathbb{P}( z_{t} ,x_{t} \mid x_{t-1} ,\pi _{t-1})\cdot \\
            &\left[ b( x_{0})\prod _{k=1}^{t-1}\mathbb{P}( z_{k} ,x_{k} \mid x_{k-1} ,\pi _{k-1}) -b( x_{0})\prod _{k=1}^{t-1}\bar{\mathbb{P}}( z_{k} ,x_{k} \mid x_{k-1} ,\pi _{k-1})\right] +\mathcal{R}_{\max} \delta _{t} \notag\\
            =&\mathcal{R}_{\max}\sum _{z_{1:t-1}}\sum _{x_{0:t-1}}\left[ b( x_{0})\prod _{k=1}^{t-1}\mathbb{P}( z_{k} ,x_{k} \mid x_{k-1} ,\pi _{k-1}) -b( x_{0})\prod _{k=1}^{t-1}\bar{\mathbb{P}}( z_{k} ,x_{k} \mid x_{k-1} ,\pi _{k-1})\right]\\ 
            &+\mathcal{R}_{\max} \delta _{t}, \notag
        \end{align}
        following similar steps recursively,
        \begin{equation}
            =\ldots=\mathcal{R}_{\max}\sum _{\tau =1}^{t} \delta _{\tau }.
        \end{equation}
        Finally, applying similar steps for every time step $t\in [1,T]$ results in,
        \begin{equation} \label{bound_delta}
            \left|V^\pi(b_t) - \bar{V}^\pi(b_t)\right| \leq \mathcal{R}_{\max}\sum _{t=1}^{T}\sum _{\tau =1}^{t} \delta _{\tau }
        \end{equation}
        where,
        \begin{align} 
            \delta_\tau &=\sum _{z_{1:\tau}}\sum _{x_{0:\tau}}\prod _{k=1}^{\tau-1}\bar{\mathbb{P}}( z_{k} ,x_{k} \mid x_{k-1} ,\pi _{k-1}) b( x_{0})[\mathbb{P}( z_{\tau} ,x_{\tau} \mid x_{\tau-1} ,\pi _{\tau-1}) -\ \bar{\mathbb{P}}( z_{\tau} ,x_{\tau} \mid x_{\tau-1} ,\pi _{\tau-1})]\notag\\
            &=\sum _{z_{1:\tau-1}}\sum _{x_{0:\tau-1}}\prod _{k=1}^{\tau-1}\bar{\mathbb{P}}( z_{k} ,x_{k} \mid x_{k-1} ,\pi _{k-1}) b( x_{0})[1-\ \sum _{z_{\tau}}\sum _{x_{\tau}}\bar{\mathbb{P}}( z_{\tau} ,x_{\tau} \mid x_{\tau-1} ,\pi _{\tau-1})] \label{delta}
        \end{align}
        plugging the term in \eqref{delta} to \eqref{bound_delta} and expanding the terms results in the desired bound,
        \begin{equation}
            \left|V^\pi(b_t) - \bar{V}^\pi(b_t)\right|\leq \! \mathcal{R}_{\max} \! \! \! \sum _{\tau=t+1}^{T} \!\! \left[ 1\! -\! \! \! \sum_{z_{t+1:{\tau}}}\sum
            _{x_{t:{\tau}}}b( x_{t}) \! \! \prod _{k=t+1}^{\tau}\bar{\mathbb{P}}( z_{k} \mid x_{k})
            \mathbb{P}( x_{k}\mid x_{k-1} ,\pi _{k-1})\right]
        \end{equation}
        which concludes our derivation.
    \end{proof}
% \end{theorem}

\subsubsection{{\Cref{lemm1}}}
% \begin{lemma}%\label{lemma1}
	The optimal value function can be bounded as
	\begin{equation}
		V^{\pi*}(b_t)  \leq    \textsc{UDB}^{\pi}(b_t),
	\end{equation}
	where the policy $\pi$ is determined according to Bellman optimality over the UDB, i.e.
	\begin{align}
		\textsc{UDB}^{\pi}(b_t) &\triangleq  \max_{a_t \in \mathcal{A}} [\bar{Q}^{\pi}(b_t, a_t) + \epsilon_z^\pi(b_t, a_t)]
		\\
		& =  \max_{a_t \in \mathcal{A}} [r(b_t, a_t) + \bar{\mathbb{E}}_{z_{t+1} | b_t,a_t}[\bar{V}^\pi(b_{t+1})] + \epsilon_z^\pi(b_t, a_t)].
	\end{align}
\label{app:proof-lemm1}
\begin{proof}% \label{proof: lemma1}
    In the following, we prove by induction that applying the Bellman optimality
    operator on upper bounds to the value function in finite-horizon POMDPs will
    result in an upper bound on the optimal value function. The notations are the
    same as the ones presented in the main body of the paper. We restate some of
    the definitions from the paper for convenience. 
    
    The policy $\pi_t(b_t)$
    determined by applying Bellman optimality at belief $b_t$, i.e.,
\begin{equation}
    \pi_t(b_t) = \arg\max_{a_t \in \mathcal{A}} [\bar{Q}^{\pi}(b_t, a_t) + \epsilon_z^\pi(b_t, a_t)].
\end{equation}
As it will be needed in the following proof, we also define the value of a
belief which includes in its history at least one observation out of the
simplified set, e.g. $\displaystyle H_{t} =\{a_{0} ,z_{1} ,\dotsc ,z_{k} \notin
\bar{\mathcal{Z}} ,\dotsc ,z_{t}\}$ as being equal to zero. Explicitly,
\begin{equation}
\bar{V}_{t}^{\pi }(\mathbb{P}( x_{t} \mid a_{0} ,z_{1} ,\dotsc ,z_{k} \notin \bar{\mathcal{Z}} ,\dotsc ,z_{t})) \equiv 0\ \ \forall k\in [ 1,t] .
\end{equation}
We also use the following simple bound,
\begin{equation}
    V_{t,\max} \triangleq \mathcal{R}_{\max} \cdot ( T-t-1)
\end{equation}

\textbf{Base case $\displaystyle ( t=T)$} - At the final time step $T$,
for each belief we set the value function to be equal to the reward value at
that belief state, $b_T$ and taking the action that maximizes the immediate
reward,
\begin{equation}
    \textsc{UDB}^{\pi }( b_{T}) =\max_{a_{T}}\{r( b_{T} ,a_{T}) +\epsilon _{z}( b_{T} ,a_{T})\} \equiv\arg\max_{a_{T}}\{r( b_{T} ,a_{T})\}
\end{equation}
which provides an upper bound for the optimal value function for the final time
step, $\displaystyle V_{T}^{\star }( b_{T}) \leq \textsc{UDB}^{\pi }( b_{T})$.\\
\textbf{Induction hypothesis} - Assume that for a given time step, $t$, for all
belief states the following holds,

\begin{equation}
    V_{t}^{\star }( b_{t}) \leq \textsc{UDB}^{\pi }( b_{t}) .
\end{equation}
\textbf{Induction step} - We will show that the hypothesis holds for time step
$t-1$. By the induction hypothesis, 

\begin{equation}
    V_{t}^{\star }( b_{t}) \leq \textsc{UDB}^{\pi }( b_{t}) \ \ \forall b_{t} ,
\end{equation}
thus, 
\begin{gather}
    Q^{\star }( b_{t-1} ,a_{t-1}) =r( b_{t-1} ,a_{t-1}) +\sum _{z_{t} \in \mathcal{Z}}\mathbb{P}\left( z_{t} \mid H_{t}^{-}\right) V_{t}^{\star }( b( z_{t}))\\
    \leq r( b_{t-1} ,a_{t-1}) +\sum _{z_{t} \in \mathcal{Z}}\mathbb{P}\left( z_{t} \mid H_{t}^{-}\right) \textsc{UDB}^{\pi }( b( z_{t})) \ \\
    =r( b_{t-1} ,a_{t-1}) +\sum _{z_{t} \in \mathcal{Z}}\mathbb{P}\left( z_{t} \mid H_{t}^{-}\right)\left[\bar{V}_{t}^{\pi }( b_{t}) +\epsilon _{z}^{\pi }( b_{t})\right].
\end{gather}
For the following transition, we make use of lemma \ref{lemma0},
\begin{gather}
    =r( b_{t-1} ,a_{t-1}) +\bar{\mathbb{E}}_{z_{t} \mid b_{t-1} ,a_{t-1}}\left[\bar{V}_{t}^{\pi }( b_{t})\right] +\epsilon _{z}^{\pi }( b_{t-1} ,a_{t-1})\\
    \equiv \textsc{UDB}^{\pi }( b_{t-1} ,a_{t-1}) .
\end{gather}
Therefore, under the induction hypothesis, $\displaystyle Q_{t-1}^{\star }(
b_{t-1} ,a_{t-1}) \leq \textsc{UDB}^{\pi }( b_{t-1} ,a_{t-1})$. Taking the maximum over
all actions $a_t$,
\begin{gather}
    \textsc{UDB}^{\pi }( b_{t-1}) =\max_{a_{t-1} \in \mathcal{A}}\left\{\textsc{UDB}^{\pi }( b_{t-1} ,a_{t-1})\right\}\\
    \geq \max_{a_{t-1} \in \mathcal{A}}\left\{Q_{t-1}^{\star }( b_{t-1} ,a_{t-1})\right\} =V_{t-1}^{\star }( b_{t-1}) , \notag
\end{gather}
which completes the induction step and the required proof.
\end{proof}
% \end{lemma}

% \subsubsection{{\Cref{lemma0}}}
\begin{lemma} \label{lemma0}
    Let $b_t$ denote a belief state and $\pi_t$ a
    policy at time $t$. Let $\bar{\mathbb{P}}(z_t \mid x_t)$ be the simplified
    observation model which represents the likelihood of observing $z_t$ given
    $x_t$. Then, the following terms are equivalent,\\
    \begin{gather}
        \mathbb{E}_{z_{t}}\left[\bar{V}_{t}^{\pi }( b_{t}) +\epsilon _{z}^{\pi }( b_{t})\right]
        = \bar{\mathbb{E}}_{z_{t}}\left[\bar{V}_{t}^{\pi }( b_{t})\right] +\epsilon _{z}^{\pi }( b_{t-1} ,a_{t-1})
        %\\ \sum _{z_{t} \in \mathcal{Z}}\mathbb{P}\left( z_{t} \mid H_{t}^{-}\right)\left[\bar{V}_{t}^{\pi }( b_{t}) +\epsilon _{z}^{\pi }( b_{t})\right] =\sum _{z_{t} \in \mathcal{Z}}\bar{\mathbb{P}}\left( z_{t} \mid H_{t}^{-}\right)\left[\bar{V}_{t}^{\pi }( b_{t})\right] +\epsilon _{z}^{\pi }( b_{t-1} ,a_{t-1})
    \end{gather}
    
\begin{proof} \label{proof:lemma0}
\begin{gather}
    \mathbb{E}_{z_{t}}\left[\bar{V}_{t}^{\pi }( b_{t}) +\epsilon _{z}^{\pi }( b_{t})\right] = \\
    %\sum _{z_{t} \in \mathcal{Z}}\mathbb{P}\left( z_{t} \mid H_{t}^{-}\right)\left[\bar{V}_{t}^{\pi }( b_{t}) +\epsilon _{z}^{\pi }( b_{t})\right] =\\
    \mathbb{E}_{z_{t}}\left[\bar{V}_{t}^{\pi }( b_{t})\right] +\mathbb{E}_{z_{t}}\left[\mathcal{R}_{\max}\sum _{\tau =t+1}^{T}\left[ 1-\sum _{z_{t+1:\tau }}\sum _{x_{t:\tau }} b_{t}\prod _{k=t+1}^{\tau }\bar{\mathbb{P}}( z_{k} \mid x_{k})\mathbb{P}( x_{k} \mid x_{k-1} ,\pi _{k-1})\right]\right]\\
     \notag
    \end{gather}
    focusing on the second summand,
\begin{gather}
    \sum _{z_{t} \in \mathcal{Z}}\mathbb{P}\left( z_{t} \mid H_{t}^{-}\right)\mathcal{R}_{\max}\sum _{\tau =t+1}^{T}\left[ 1-\sum _{z_{t+1:\tau }}\sum _{x_{t:\tau }} b_{t}\prod _{k=t+1}^{\tau }\bar{\mathbb{P}}( z_{k} \mid x_{k})\mathbb{P}( x_{k} \mid x_{k-1} ,\pi _{k-1})\right]\\
    =\mathcal{R}_{\max}\sum _{\tau =t+1}^{T}\left[ 1-\sum _{z_{t}}\mathbb{P}\left( z_{t} \mid H_{t}^{-}\right)\sum _{z_{t+1:\tau }}\sum _{x_{t:\tau }} b( x_{t})\prod _{k=t+1}^{\tau }\bar{\mathbb{P}}( z_{k} \mid x_{k})\mathbb{P}( x_{k} \mid x_{k-1} ,\pi _{k-1})\right]
    \end{gather}
    by marginalizing over $x_{t-1}$,

\begin{gather}
    =\mathcal{R}_{\max}\sum _{\tau =t+1}^{T}[ 1-\sum _{z_{t}}\mathbb{P}\left( z_{t} \mid H_{t}^{-}\right)\sum _{z_{t+1:\tau }}\sum _{x_{t-1:\tau }}\frac{\bar{\mathbb{P}}( z_{t} \mid x_{t})\mathbb{P}( x_{t} \mid x_{t-1} ,\pi _{t-1}) b( x_{t-1})}{\mathbb{P}\left( z_{t} \mid H_{t}^{-}\right)} \cdot \\
    \prod _{k=t+1}^{\tau }\bar{\mathbb{P}}( z_{k} \mid x_{k})\mathbb{P}( x_{k} \mid x_{k-1} ,\pi _{k-1})] \notag
    \end{gather}
    canceling out the denominator,

\begin{gather}
    =\mathcal{R}_{\max}\sum _{\tau =t+1}^{T}[ 1-\sum _{z_{t:\tau }}\sum _{x_{t-1:\tau }}\bar{\mathbb{P}}( z_{t} \mid x_{t})\mathbb{P}( x_{t} \mid x_{t-1} ,a_{t-1}) b( x_{t-1}) \cdot \\
    \prod _{k=t+1}^{\tau }\bar{\mathbb{P}}( z_{k} \mid x_{k})\mathbb{P}( x_{k} \mid x_{k-1} ,\pi _{k-1})] \equiv \epsilon _{z}^{\pi }( b_{t-1} ,a_{t-1}) \notag
    \end{gather}
    it is left to show that $\mathbb{E}_{z_{t} \mid b_{t-1}
    ,a_{t-1}}\left[\bar{V}_{t}^{\pi }( b_{t})\right]
    =\bar{\mathbb{E}}_{z_{t} \mid b_{t-1}
    ,a_{t-1}}\left[\bar{V}_{t}^{\pi }( b_{t})\right]$. By the definition of
    a value function of a belief not included in the simplified set, we have
    that,

\begin{align}
    \mathbb{E}_{z_{t} \mid b_{t-1}
    ,a_{t-1}}\left[\bar{V}_{t}^{\pi }( b_{t})\right]
    &=\sum _{z_{t} \in \mathcal{Z}}\mathbb{P}\left( z_{t} \mid H_{t}^{-}\right)\bar{V}_{t}^{\pi }( b_{t})\\
    &=\sum _{z_{t} \in \bar{\mathcal{Z}}}\mathbb{P}\left( z_{t} \mid H_{t}^{-}\right)\bar{V}_{t}^{\pi }( b_{t}) +\sum _{z_{t} \in \mathcal{Z} \backslash \bar{\mathcal{Z}}}\mathbb{P}\left( z_{t} \mid H_{t}^{-}\right)\bar{V}_{t}^{\pi }( b_{t})\\
    &=\sum _{z_{t} \in \bar{\mathcal{Z}}}\bar{\mathbb{P}}\left( z_{t} \mid H_{t}^{-}\right) \cdot \bar{V}_{t}^{\pi }( b_{t}) +\sum _{z_{t} \in \mathcal{Z} \backslash \bar{\mathcal{Z}}}\mathbb{P}\left( z_{t} \mid H_{t}^{-}\right) \cdot 0\\
    &=\bar{\mathbb{E}}_{z_{t} \mid b_{t-1} ,a_{t-1}}\left[\bar{V}_{t}^{\pi }( b_{t})\right] ,
\end{align}
    which concludes the derivation.
    \end{proof}
\end{lemma}

% \subsection{Corollary 1.1}
\subsubsection{{\Cref{cor:1}}}
We restate the definition of UDB exploration criteria,
\begin{equation}%\label{eq:exploration}
	a_t = \arg \max_{a_t \in \mathcal{A}}[   \textsc{UDB}^{\pi}(b_t,a_t)] =
	\arg \max_{a_t \in \mathcal{A}} [\bar{Q}^{\pi}(b_t, a_t) + \epsilon_z^\pi(b_t, a_t)].
\end{equation}
\begin{corollary}
    Using Lemma \ref{lemm1} and the exploration criteria defined in
    \eqref{eq:exploration} guarantees convergence to the optimal value
    function.
\begin{proof} \label{proof:corollary1.1}
    Let us define a sequence of bounds, $\textsc{UDB}_n^{\pi}(b_t)$ and a
    corresponding difference value between $\textsc{UDB}_n$ and the simplified
    value function,
    \begin{equation}
        \textsc{UDB}_n^{\pi}(b_t) - \bar{V}_n^{\pi}(b_t) = \epsilon_{n,z}^\pi(b_t),
    \end{equation} 
    where $n\in[0, \left|\mathcal{Z}\right|]$ corresponds to the number of
    unique observation instances within the simplified observation set,
    $\bar{\mathcal{Z}}_n$, and $\left|\mathcal{Z}\right|$ denotes the
    cardinality of the complete observation space. Additionally, for the clarity
    of the proof and notations, assume that by construction the simplified set
    is chosen such that
    $\bar{\mathcal{Z}}_n(H_t)\equiv\bar{\mathcal{Z}}_n$ remains
    identical for all time steps $t$ and history sequences, $H_t$ given $n$. By
    the definition of $\epsilon_{n,z}^\pi(b_t)$, 
    \begin{equation}
        \epsilon_{n,z}^\pi(b_t)= \mathcal{R}_{\max} \! \! \! \sum _{\tau=t+1}^{T} \!\! \left[ 1\! -\! \! \! \sum_{z_{t+1:{\tau}}\in \bar{\mathcal{Z}}_n}\sum
        _{x_{t:{\tau}}}b( x_{t}) \! \! \prod _{k=t+1}^{\tau}\bar{\mathbb{P}}( z_{k} \mid x_{k})
        \mathbb{P}( x_{k}\mid x_{k-1} ,\pi _{k-1})\right],
    \end{equation}
    we have that $\epsilon_{n,z}^\pi(b_t)\rightarrow 0$ as
    $n\rightarrow\left|\mathcal{Z}\right|$, since 
    \begin{equation}
        \sum_{z_{t+1:{\tau}}\in\bar{\mathcal{Z}}_n}\sum
    _{x_{t:{\tau}}}b( x_{t}) \prod _{k=t+1}^{\tau}\bar{\mathbb{P}}( z_{k} \mid x_{k})
    \mathbb{P}( x_{k}\mid x_{k-1} ,\pi _{k-1}) \rightarrow 1
    \end{equation}
    as more unique observation elements are added to the simplified observation
    space, $\bar{\mathcal{Z}}_n$, eventually recovering the entire support
    of the discrete observation distribution. 
    % Additionally, note that
    % $\epsilon_{n,z}^\pi(b_t)$ is monotonically decreasing with growing number of
    % elements in $\bar{\mathcal{Z}}_n$, although we do not formally prove it
    % here.

    From lemma \ref{lemm1} we have that, for all $n\in[0,
    \left|\mathcal{Z}\right|]$ the following holds,
	\begin{equation}
		V^{\pi*}(b_t) \leq \textsc{UDB}_n^{\pi}(b_t) = \bar{V}_n^{\pi}(b_t) + \epsilon_{n,z}^\pi(b_t).
	\end{equation}
    Additionally, from theorem \ref{thm:valueFuncSimplifiedObs} we have that,
    \begin{equation}
        \left|V^\pi(b_t) - \bar{V}_n^\pi(b_t)\right|\leq \epsilon_{n,z}^\pi(b_t),
    \end{equation}
    for any policy $\pi$ and subset
    $\bar{\mathcal{Z}}_n\subseteq\mathcal{Z}$, thus,
    \begin{equation}
        \bar{V}_n^\pi(b_t) - \epsilon_{n,z}^\pi(b_t) \leq V^\pi(b_t) \leq V^{\pi*}(b_t) \leq \bar{V}_n^{\pi}(b_t) + \epsilon_{n,z}^\pi(b_t).
    \end{equation}

    Since $\epsilon_{n,z}^\pi(b_t)\rightarrow 0$ as
    $n\rightarrow\left|\mathcal{Z}\right|$, and $\left|\mathcal{Z}\right|$
    is finite, it is guaranteed that $\textsc{UDB}_n^{\pi}(b_t)
    \xrightarrow{n\rightarrow \left|\mathcal{Z}\right|} V^{\pi*}(b_t)$ which
    completes our proof.
\end{proof}
\end{corollary}
Moreover, depending on the algorithm implementation, the number of iterations
can be finite (e.g. by directly choosing actions and observations to minimize the
bound). A stopping criteria can also be verified by calculating the difference
between the upper and lower bounds. The optimal solution is obtained once the
upper bound equals the lower bound.
    
\subsubsection{{\Cref{thm:valueFuncSimplifiedBoth}}}
% \begin{theorem*}
    Let $b_t$ belief state at time $t$, and $T$ be the last time step of the POMDP.
    Let $V^{\pi}(b_t)$ be the theoretical value function by following a policy
    $\pi$, and let $\bar{V}^{\pi}(b_t)$ be the simplified value function, 
    as defined in \eqref{def:simplifiedValueFunc}, by following the same policy.
    Then, for any policy $\pi$, the difference between the theoretical and
    simplified value functions is bounded as follows,
    \begin{equation}
        \left|V^\pi(b_t) \! - \! \bar{V}^\pi(b_t)\right|\leq \! \mathcal{R}_{\max} \! \! \! \sum _{\tau=t+1}^{T} \!\! \left[ 1\! -\! \! \! \sum_{z_{t+1:{\tau}}}\sum
        _{x_{t:{\tau}}}b( x_{t}) \! \! \prod _{k=t+1}^{\tau}\bar{\mathbb{P}}( z_{k} \mid x_{k})
        \mathbb{P}( x_{k}\mid x_{k-1} ,\pi _{k-1})\right]\triangleq \epsilon^\pi(b_t).
    \end{equation}
% \end{theorem*}
\begin{proof} \label{proof:thm2}
Recall that we define $ \tau _{t} =\{x_{0} ,a_{0} ,z_{1} ,x_{1} ,a_{1} ,\dotsc
,a_{T-1} ,x_{t} ,z_{t}\}$. Then the value function is defined as, 
\begin{equation}
    V^{\pi }( b_{0}) =\sum _{\tau _{T}}\mathbb{P}^{\pi }( \tau _{T})\left[\sum _{t=0}^{T} r( x_{t} ,a_{t})\right]
\end{equation}
applying chain rule and rearranging terms,

\begin{gather}
    =\sum _{\tau _{T}}\mathbb{P}^{\pi }( x_{1:T} ,z_{1:T} ,a_{1:T} \mid \tau _{0})\mathbb{P}^{\pi }( \tau _{0})\left[\sum _{t=0}^{T} r( x_{t} ,a_{t})\right]\\
    =\sum _{\tau _{0}}\mathbb{P}^{\pi }( \tau _{0})\sum _{x_{1:T} ,z_{1:T} ,a_{1:T}}\mathbb{P}^{\pi }( x_{1:T} ,z_{1:T} ,a_{1:T} \mid \tau _{0})\left[\sum _{t=0}^{T} r( x_{t} ,a_{t})\right]\\
    =\sum _{\tau _{0}}\mathbb{P}^{\pi }( \tau _{0})\left[ r( x_{0} ,a_{0}) +\sum _{x_{1:T} ,z_{1:T} ,a_{1:T}}\mathbb{P}^{\pi }( x_{1:T} ,z_{1:T} ,a_{1:T} \mid \tau _{0})\left[\sum _{t=1}^{T} r( x_{t} ,a_{t})\right]\right]
\end{gather}
nullifying instances of the complete probability distribution,
$ \mathbb{P}^{\pi }( \cdot )$, is denoted as a simplified distribution,
$ \bar{\mathbb{P}}^{\pi }( \cdot )$. We can then split and bound from
above the value function, such that the simplified value function consideres
only a subset of the trajectories at time $t=0$,

\begin{gather}
    \leq \sum _{\tau _{0}}\bar{\mathbb{P}}^{\pi }( \tau _{0})\left[ r( x_{0} ,a_{0}) +\sum _{x_{1:T} ,z_{1:T} ,a_{1:T}}\mathbb{P}^{\pi }( x_{1:T} ,z_{1:T} ,a_{1:T} \mid \tau _{0})\left[\sum _{t=1}^{T} r( x_{t} ,a_{t})\right]\right]\\
    +\left[ 1-\sum _{\tau _{0}}\bar{\mathbb{P}}^{\pi }( \tau _{0})\right] \mathcal{V}_{max,0}
\end{gather}
We then apply similar steps on the next time step, $t=1$,

\begin{align}
    =&\sum _{\tau _{0}}\bar{\mathbb{P}}^{\pi }( \tau _{0})\left[ r( x_{0} ,a_{0}) +\sum _{x_{1:T} ,z_{1:T} ,a_{1:T}}\mathbb{P}^{\pi }( x_{2:T} ,z_{2:T} ,a_{2:T} \mid \tau _{1})\mathbb{P}^{\pi }( x_{1} ,z_{1} ,a_{1} \mid \tau _{0})\left[\sum _{t=1}^{T} r( x_{t} ,a_{t})\right]\right]\\
    &+\left[ 1-\sum _{\tau _{0}}\bar{\mathbb{P}}^{\pi }( \tau _{0})\right] \mathcal{V}_{max, 0}\\
    =&\sum _{\tau _{0}}\bar{\mathbb{P}}^{\pi }( \tau _{0})\Biggl[ r( x_{0} ,a_{0}) \\ \notag
    &+\sum _{x_{1} ,z_{1} ,a_{1}}\mathbb{P}^{\pi }( x_{1} ,z_{1} ,a_{1} \mid \tau _{0})\sum _{x_{2:T} ,z_{2:T} ,a_{2:T}}\mathbb{P}^{\pi }( x_{2:T} ,z_{2:T} ,a_{2:T} \mid \tau _{1})\left[\sum _{t=1}^{T} r( x_{t} ,a_{t})\right]\Biggr]\\
    &+\left[ 1-\sum _{\tau _{0}}\bar{\mathbb{P}}^{\pi }( \tau _{0})\right] \mathcal{V}_{max, 0}\\
    =&\sum _{\tau _{0}}\bar{\mathbb{P}}^{\pi }( \tau _{0})\Biggl[ r( x_{0} ,a_{0})  \\ \notag
    &+\sum _{x_{1} ,z_{1} ,a_{1}}\mathbb{P}^{\pi }( x_{1} ,z_{1} ,a_{1} \mid \tau _{0})\Biggl[ r( x_{1} ,a_{1})  \\ \notag
    &+\sum _{x_{2:T} ,z_{2:T} ,a_{2:T}}\mathbb{P}^{\pi }( x_{2:T} ,z_{2:T} ,a_{2:T} \mid \tau _{1})\left[\sum _{t=2}^{T} r( x_{t} ,a_{t})\right]\Biggr]\Biggr]\\ \notag
    &+\left[ 1-\sum _{\tau _{0}}\bar{\mathbb{P}}^{\pi }( \tau _{0})\right] \mathcal{V}_{max, 0}
\end{align}
\begin{align}
    \leq& \sum _{\tau _{0}}\bar{\mathbb{P}}^{\pi }( \tau _{0})\Biggl[ r( x_{0} ,a_{0})   \\ \notag 
    &+\sum _{x_{1} ,z_{1} ,a_{1}}\bar{\mathbb{P}}^{\pi }( x_{1} ,z_{1} ,a_{1} \mid \tau _{0})\Biggl[ r( x_{1} ,a_{1})   \\ \notag
    &+\sum _{x_{2:T} ,z_{2:T} ,a_{2:T}}\mathbb{P}^{\pi }( x_{2:T} ,z_{2:T} ,a_{2:T} \mid \tau _{1})\left[\sum _{t=2}^{T} r( x_{t} ,a_{t})\right]\Biggr]\Biggr]\\
    &+\sum _{\tau _{0}}\bar{\mathbb{P}}^{\pi }( \tau _{0})\left[ 1-\sum _{x_{1} ,z_{1} ,a_{1}}\bar{\mathbb{P}}^{\pi }( x_{1} ,z_{1} ,a_{1} \mid \tau _{0})\right] \mathcal{V}_{max, 1} +\left[ 1-\sum _{\tau _{0}}\bar{\mathbb{P}}^{\pi }( \tau _{0})\right] \mathcal{V}_{max, 0}
\end{align}
which results in,

\begin{gather}
    =\sum _{\tau _{0}}\bar{\mathbb{P}}^{\pi }( \tau _{0})\Biggl[ r( x_{0} ,a_{0})  \\ \notag 
    +\sum _{x_{1} ,z_{1} ,a_{1}}\bar{\mathbb{P}}^{\pi }( x_{1} ,z_{1} ,a_{1} \mid x_{0} ,a_{0})\left[ r( x_{1} ,a_{1}) +\sum _{x_{2:T} ,z_{2:T} ,a_{2:T}}\mathbb{P}^{\pi }( x_{2:T} ,z_{2:T} ,a_{2:T} \mid \tau _{1})\left[\sum _{t=2}^{T} r( x_{t} ,a_{t})\right]\right]\Biggr]\\
    +\left[\sum _{\tau _{0}}\bar{\mathbb{P}}^{\pi }( \tau _{0}) -\sum _{\tau _{1}}\bar{\mathbb{P}}^{\pi }( \tau _{1})\right] \mathcal{V}_{max, 1} +\left[ 1-\sum _{\tau _{0}}\bar{\mathbb{P}}^{\pi }( \tau _{0})\right] \mathcal{V}_{max, 0}
\end{gather}
Performing the same steps iteratively up to time $t=T$, yields the desired
outcome,
\begin{equation}
    V^{\pi }( b_{0}) \leq \sum _{t=0}^{T}\sum _{\tau _{t}}\bar{\mathbb{P}}^{\pi }( \tau _{t}) r( x_{t} ,a_{t}) +\mathcal{V}_{max, 0}\left[ 1-\sum _{\tau _{0}}\bar{\mathbb{P}}^{\pi }( \tau _{0})\right] +\sum _{t=0}^{T-1} \mathcal{V}_{max, t+1}\left[\sum _{\tau _{t}}\bar{\mathbb{P}}^{\pi }( \tau _{t}) -\sum _{\tau _{t+1}}\bar{\mathbb{P}}^{\pi }( \tau _{t+1})\right]
\end{equation}
\end{proof}

% \subsection[short]{Optimality Guarantees} 

\subsubsection{{\Cref{lemma:udbOptimality}}}
    Let $\mathcal{A}$ be the set of
    actions and $\mathcal{U}_0^\star(H_t)$, $\mathcal{L}_0^\star(H_t)$ be the upper
    and lower bounds of node $H_t$ chosen according to,

    % \begin{align} \label{def:optimalRecursive}
    %     &\mathcal{U}_0^\star(H_t) \triangleq \max_{a_t} \sum_{\tau_{t}\in \mathcal{T}(H_t)} \bar{\mathbb{P}}(\tau_t) r(x_t, a_t) + \sum_{z_{t+1}\in \bar{\mathcal{Z}}( H_{t},a_t)}\left[\mathcal{U}_0^\star(H_{t+1}) + \mathcal{V}_{\max, t} \left(\sum_{\tau_t\in \mathcal{T}(H_t)}\bar{\mathbb{P}}(\tau_t) - \sum_{\tau_{t+1}\in \mathcal{T}(H_{t+1})}\bar{\mathbb{P}}(\tau_{t+1})\right)\right]\\
    %     &\mathcal{L}_0^\star(H_t) \triangleq \max_{a_t} \sum_{\tau_{t}\in \mathcal{T}(H_t)} \bar{\mathbb{P}}(\tau_t) r(x_t, a_t) + \sum_{z_{t+1}\in \bar{\mathcal{Z}}( H_{t},a_t)}\left[\mathcal{L}_0^\star(H_{t+1}) - \mathcal{V}_{\max, t} \left(\sum_{\tau_t\in \mathcal{T}(H_t)}\bar{\mathbb{P}}(\tau_t) - \sum_{\tau_{t+1}\in \mathcal{T}(H_{t+1})}\bar{\mathbb{P}}(\tau_{t+1})\right)\right],
    % \end{align}
    \begin{align}
        &\mathcal{U}^\star_0(H_t) \triangleq \sum_{\tau_{t}\in \mathcal{T}(H_t)} \bar{\mathbb{P}}(\tau_t) \left[r(x_t, a_t) + \mathcal{V}_{\max, t}\right]+ \sum_{z_{t+1}\in \bar{\mathcal{Z}}(H_t, a_t)}\left[\mathcal{U}^\star_0(H_{t+1}) - \sum_{\tau_{t+1}\in \mathcal{T}(H_{t+1})}\!\!\!\!\!\!\!\!\!\bar{\mathbb{P}}(\tau_{t+1})\mathcal{V}_{\max, t}\right]\\
        &\mathcal{L}^\star_0(H_t) \triangleq \sum_{\tau_{t}\in \mathcal{T}(H_t)} \bar{\mathbb{P}}(\tau_t) \left[r(x_t, a_t) + \mathcal{V}_{\min, t}\right]+ \sum_{z_{t+1}\in \bar{\mathcal{Z}}(H_t, a_t)}\left[\mathcal{L}^\star_0(H_{t+1}) - \sum_{\tau_{t+1}\in \mathcal{T}(H_{t+1})}\!\!\!\!\!\!\!\!\!\bar{\mathbb{P}}(\tau_{t+1})\mathcal{V}_{\min, t}\right]
    \end{align}
    and,
    \begin{align}
        &\mathcal{U}_0^\star(H_T) \triangleq \sum_{\tau_{T}\in \mathcal{T}(H_T)} \bar{\mathbb{P}}(\tau_T) r(x_T),
        &\mathcal{L}_0^\star(H_T) \triangleq \sum_{\tau_{T}\in \mathcal{T}(H_T)} \bar{\mathbb{P}}(\tau_T) r(x_T).
    \end{align}

    Then, the optimal root-value is bounded by,
    \begin{equation}
    \mathcal{L}_0^\star(H_0) \leq V^{\pi^*}(H_0) \leq \mathcal{U}_0^\star(H_0).
    \end{equation}

\begin{proof} \label{proof:optimalRecursive}
    We wish to show that $ \mathcal{L}_{0}^{\star }( H_{0}) \leq V^{\pi ^{*}}(
        b_{0}) \leq \mathcal{U}_{0}^{\star }( H_{0})$. We derive a proof for one
        side of the inequality, while the other follows similarly. First note
        that,
    
    \begin{equation}
        V^{\pi ^{*}}( b_{0}) \leq \mathcal{U}_{0}^{\pi ^{*}}( H_{0}) \leq \underset{\pi \in \Pi }{\max}\mathcal{U}_{0}^{\pi }( H_{0})
    \end{equation}
    where the first inequality is due to Theorem
    \ref{thm:valueFuncSimplifiedBoth}, and the second inequality is true by
    definition. However, the claim in Corollary \ref{lemma:udbOptimality} is a
    recursive claim, while the bound provided in Theorem
    \ref{thm:valueFuncSimplifiedBoth} only holds with respect to the root. Thus,
    for completeness, we also need to show that the best action can be chosen
    recursively, even though the bound is `partial` in different parts of the
    tree.

    \begin{align*}
        &\underset{\pi _{0:T} \in \Pi }{\max}\mathcal{U}_{0}^{\pi }( H_{0})\\
        &=\underset{\pi _{0:T} \in \Pi }{\max}\sum _{\tau _{0} \in \mathcal{T}( H_{0})}\overline{\mathbb{P}}( \tau _{0})[ r( x_{0} ,\pi _{0}) +\mathcal{V}_{max,0}] +\sum _{z_{1} \in \overline{\mathcal{Z}}( H_{0} ,\pi _{0})}\left[\mathcal{U}_{0}^{\pi }( H_{1}) -\sum _{\tau _{1} \in \mathcal{T}( H_{1})}\overline{\mathbb{P}}( \tau _{1}) \mathcal{V}_{max,0}\right]\\
        &=\underset{\pi _{0} \in \Pi }{\max}\left\{\sum _{\tau _{0} \in \mathcal{T}( H_{0})}\overline{\mathbb{P}}( \tau _{0})[ r( x_{0} ,\pi _{0}) +\mathcal{V}_{max,0}] +\underset{\pi _{1:T} \in \Pi }{\max}\sum _{z_{1} \in \overline{\mathcal{Z}}( H_{0} ,\pi _{0})}\left[\mathcal{U}_{0}^{\pi }( H_{1}) -\sum _{\tau _{1} \in \mathcal{T}( H_{1})}\overline{\mathbb{P}}( \tau _{1}) \mathcal{V}_{max,0}\right]\right\}\\
        &=\underset{a_{0}}{\max}\left\{\sum _{\tau _{0} \in \mathcal{T}( H_{0})}\overline{\mathbb{P}}( \tau _{0})[ r( x_{0} ,a_{0}) +\mathcal{V}_{max,0}] +\sum _{z_{1} \in \overline{\mathcal{Z}}( H_{0} ,a_{0})}\left[\underset{\pi _{1:T} \in \Pi }{\max} \mathcal{U}_{0}^{\pi }( H_{1}) -\sum _{\tau _{1} \in \mathcal{T}( H_{1})}\overline{\mathbb{P}}( \tau _{1}) \mathcal{V}_{max,0}\right]\right\}
    \end{align*}
    which continues similarly up to time $t=T$, which completes the proof,

    \begin{equation}
        V^{\pi ^{*}}( b_{0}) \leq \mathcal{U}_{0}^{\pi ^{*}}( H_{0}) \leq \underset{\pi \in \Pi }{\max}\mathcal{U}_{0}^{\pi }( H_{0}) = \mathcal{U}_{0}^{\star }( H_{0}). \qed
    \end{equation}
\end{proof}

\subsubsection{{\Cref{lemma:convergenceToOptimal}}}
    Performing exploration based on
    \eqref{eq:rootBoundExplor}, \eqref{eq:obsExplor} and \eqref{eq:stateExplor}
    ensures that the algorithm converges to the optimal value function within a
    finite number of planning iterations.
\begin{proof} \label{proof:convergenceToOptimal}
    Consider a given policy $\pi$. We claim that following the state and
    observation selection criteria in equations \eqref{eq:obsExplor} and
    \eqref{eq:stateExplor} will lead to visiting unexplored trajectories
    $\tau_T$ at every iteration unless all relevant trajectories have already
    been explored.

    To show this, note that the upper bound $\mathcal{U}_0^{\star}((H_t, a_t,
    o_{t+1}))$ and the lower bound $\mathcal{L}_0^{\star}((H_t, a_t, o_{t+1}))$ will
    converge when the bound interval is zero, i.e.,
    \begin{align}
        \mathcal{U}_0^{\star}((H_t, a_t, o_{t+1})) - \mathcal{L}_0^{\star}((H_t, a_t, o_{t+1})) = 0.
    \end{align}
    This convergence occurs when all future trajectories by following policy
    $\pi$ from node $H_{t+1}=(H_{t}, a_t, o_{t+1})$ until the end of the horizon
    were explored,
    \begin{align*}
        &\mathcal{U}_0^\pi(H_{t+1}) - \mathcal{L}_0^\pi(H_{t+1}) = \\
        &=\sum_{\tau_{t+1} \in \mathcal{T}(H_{t+1})}\bar{\mathbb{P}}(\tau_{t+1})\mathcal{V}_{\max, {t+1}} + \sum_{z_{t+2}\in \bar{\mathcal{Z}}(H_{t+1}, \pi_{t+1})}\left[\mathcal{U}^\pi_0(H_{t+2}) - \!\!\!\!\!\!\!\!\!\sum_{\tau_{t+2}\in \mathcal{T}(H_{t+2})}\!\!\!\!\!\!\!\!\!\bar{\mathbb{P}}(\tau_{t+2})\mathcal{V}_{\max, {t+1}}\right]\\
        &- \left[\sum_{\tau_{t+1} \in \mathcal{T}(H_{t+1})}\bar{\mathbb{P}}(\tau_{t+1})\mathcal{V}_{\min, {t+1}} + \sum_{z_{t+2}\in \bar{\mathcal{Z}}(H_{t+1}, \pi_{t+1})}\left[\mathcal{L}^\pi_0(H_{t+2}) - \!\!\!\!\!\!\!\!\!\sum_{\tau_{t+2}\in \mathcal{T}(H_{t+2})}\!\!\!\!\!\!\!\!\!\bar{\mathbb{P}}(\tau_{t+2})\mathcal{V}_{\min, t+1}\right]\right]\\
        &=\left[\sum_{\tau_{t+1} \in \mathcal{T}(H_{t+1})}\bar{\mathbb{P}}(\tau_{t+1}) - \sum_{z_{t+2}\in \bar{\mathcal{Z}}(H_{t+1}, \pi_{t+1})} \sum_{\tau_{t+2}\in \mathcal{T}(H_{t+2})}\bar{\mathbb{P}}(\tau_{t+2}) \right] \left(\mathcal{V}_{\max, {t+1}} - \mathcal{V}_{\min, {t+1}}\right) \\
        &+ \sum_{z_{t+2}\in \bar{\mathcal{Z}}(H_{t+1}, \pi_{t+1})} \left[\mathcal{U}^\pi_0(H_{t+2}) - \mathcal{L}^\pi_0(H_{t+2}) \right] 
    \end{align*}
    since $\forall t\in \left[0, T-1 \right], \mathcal{V}_{\max, {t+1}} - \mathcal{V}_{\min, {t+1}} \neq 0$, then
    $\mathcal{U}_0^\pi(H_{t+1}) - \mathcal{L}_0^\pi(H_{t+1}) = 0$ only if ,
    \begin{equation}
         \ \sum_{\tau_{t+1} \in \mathcal{T}(H_{t+1})}\bar{\mathbb{P}}(\tau_{t+1}) - \sum_{z_{t+2}\in \bar{\mathcal{Z}}(H_{t+1}, \pi_{t+1})} \sum_{\tau_{t+2}\in \mathcal{T}(H_{t+2})}\bar{\mathbb{P}}(\tau_{t+2}) = 0, \ \forall t\in \left[0, T-2 \right].
    \end{equation}
    Thus, all the simplified probability terms in the policy tree converge to 1.
    Similarly, the probability gap,
    \begin{align}
        1 - \sum_{\tau_{T}} \bar{\mathbb{P}}^{\star}(\tau_T \mid \tau_t, a_t, z_{t+1}, x)=0
    \end{align}
    only when all non-zero future trajectories with a prefix $(\tau_t, a_t, z_{t+1},
    x)$ have been explored. Finally, we are left to show that selecting actions
    based on the criteria shown in \eqref{eq:rootBoundExplor}, results in the
    optimal action upon convergence. Utilizing corollary \ref{lemma:udbOptimality},
    the proof follows similarly to the one shown in \eqref{proof:corollary1.1},
    which concludes our derivation.
\end{proof}

\VI{
\subsubsection[short]{Gap between the upper and lower bounds.} \label{lower_upper_gap}
Define
\[
\Delta(H_t)\;\triangleq\;\mathcal{U}^\pi_0(H_t)-\mathcal{L}^\pi_0(H_t),\qquad
\Delta V(H_t)\;\triangleq\;\mathcal{V}_{\max}(H_t)-\mathcal{V}_{\min}(H_t).
\]

\textbf{Step 1:  Recursion.}  
Subtract the two expressions in \eqref{def:RecursiveBound}:

\begin{align}
\Delta(H_t)
  &=\sum_{\tau_t\in\mathcal{T}(H_t)}
        \bar{\mathbb{P}}(\tau_t)\,\Delta V(H_t)
    +\!\!\!\sum_{z_{t+1}\in\bar{\mathcal{Z}}(H_t,\pi_t)}\!\!\!
       \bigl[
         \Delta(H_{t+1})
        -\!\!\!\!\sum_{\tau_{t+1}\in\mathcal{T}(H_{t+1})}\!\!\!
           \bar{\mathbb{P}}(\tau_{t+1})\,\Delta V(H_t)\bigr] .
    \label{eq:gap_recursion}
\end{align}

Because $\mathcal{T}(H_{t+1})$ contains \emph{only} those trajectories whose
next observation is \emph{inside} $\bar{\mathcal{Z}}$ \emph{and} whose next
state is \emph{inside} the current simplified state set
$\bar{\mathcal{X}}(H_{t+1})$, the probability that \emph{drops out} between the
two sums in \eqref{eq:gap_recursion} is precisely the mass of trajectories that
\emph{leave} \emph{either} simplified space at step $t+1$.

\textbf{Step 2:  Coverage-gap term.}  
Let
\[
\delta(H_t)\;\triangleq\;
\sum_{\tau_t\in\mathcal{T}(H_t)}\bar{\mathbb{P}}(\tau_t)\;-\!\!\!
\sum_{z_{t+1}\in\bar{\mathcal{Z}}(H_t,\pi_t)}\;
      \sum_{\tau_{t+1}\in\mathcal{T}(H_{t+1})}
        \bar{\mathbb{P}}(\tau_{t+1}),
\]
i.e.\ the probability that either  
(i) $z_{t+1}\notin\bar{\mathcal{Z}}(H_t,\pi_t)$  
\emph{or}  
(ii) $z_{t+1}\in\bar{\mathcal{Z}}$ but $x_{t+1}\notin\bar{\mathcal{X}}(H_{t+1})$
given $\tau_t$.
Then \eqref{eq:gap_recursion} rewrites as
\[
\Delta(H_t)\;=\;\Delta V(H_t)\,\delta(H_t)
               +\!\!\!\sum_{z_{t+1}\in\bar{\mathcal{Z}}(H_t,\pi_t)}\!\!\!\Delta(H_{t+1}),
\]
or,
\[
  \mathcal{U}^\pi_0(H_t)-\mathcal{L}^\pi_0(H_t)\;=\;
  \Delta V(H_t)\,\delta(H_t)
               +\!\!\!\sum_{z_{t+1}\in\bar{\mathcal{Z}}(H_t,\pi_t)}\!\!\!\left[\mathcal{U}^\pi_0(H_{t+1})-\mathcal{L}^\pi_0(H_{t+1})\right].
\]

% \textbf{Step 3:  Closed form.}  
% Unwinding the recursion along any trajectory
% $\tau=(x_{t:T},z_{t+1:T})$ produces

% \[
% \mathcal{U}^\pi_0(H_t)-\mathcal{L}^\pi_0(H_t)\;=\;
% \mathbb{E}_{\tau}
% \Bigl[\;
%    \sum_{k=t}^{T-1}
%         \Delta V(H_k)\,
%         \mathbf{1}\!\Bigl\{
%           \tau_{k+1}\notin\bar{\mathcal{T}}(H_{k+1})
%         \Bigr\}
%    \;\Bigm|\;H_t
% \Bigr]\!.
% \]

Hence the upper-lower gap at node $H_t$ accumulates the local value spread
$\Delta V(H_k)$ \emph{only} on steps where the trajectory exits \emph{either}
simplified space.  If all future observations and states remain in
$\bar{\mathcal{Z}}$ and $\bar{\mathcal{X}}$, the indicator is zero and the gap
collapses; conversely, frequent exits slow down the tightening of the bounds.
}

\section{Experiments}

\subsection{POMDP scenarios}
We begin with a brief description of the Partially Observable Markov Decision
Process (POMDP) scenarios implemented for the experiments. each scenario was
bounded by a finite number of time steps used for every episode, where each
action taken by the agent led to a decrement in the number of time steps left.
After the allowable time steps ended, the simulation was reset to its initial
state.

\subsubsection{Tiger POMDP}
The Tiger is a classic POMDP problem \cite{Kaelbling98ai}, involves an agent making
decisions between two doors, one concealing a tiger and the other a reward. The
agent needs to choose among three actions, either open each one of the doors or
listen to receive an observation about the tiger position. In our experiments,
the POMDP was limited horizon of 5 steps. The problem consists of 3 actions, 2
observations and 2 states.

\subsubsection{Discrete Light Dark} Is an adaptation from \cite{Sunberg18icaps}. In
this setting the agent needs to travel on a 1D grid to reach a target location.
The grid is divided into a dark region, which offers noisy observations, and a
light region, which offers accurate localization observations. The agent
receives a penalty for every step and a reward for reaching the target
location. The key challenge is to balance between information gathering by
traveling towards the light area, and moving towards the goal region.

\subsubsection{Laser Tag POMDP}
In the Laser Tag problem, \cite{Somani13nips}, an agent has to navigate through
a grid world, shoot and tag opponents by using a laser gun. The main goal is to
tag as many opponents as possible within a given time frame. The grid is
segmented into various sections that have varying visibility, characterized by
obstacles that block the line of sight, and open areas. There are five possible
actions, moving in four cardinal directions (North, South, East, West) and
shooting the laser. The observation space cardinality is
$\left|\mathcal{Z}\right|\approx 1.5 \times 10^6$, which is described as a
discretized normal distribution and reflect the distance measured by the laser.
The states reflect the agent's current position and the opponents' positions.
The agent receives a reward for tagging an opponent and a penalty for every
movement, encouraging the agent to make strategic moves and shots.

\subsubsection{Baby POMDP}
The Baby POMDP is a classic problem that represents the scenario of a baby and a
caregiver. The agent, playing the role of the caregiver, needs to infer the
baby's needs based on its state, which can be either crying or quiet. The states
in this problem represent the baby's needs, which could be hunger, discomfort or
no need. The agent has three actions to choose from: feeding, changing the
diaper, or doing nothing. The observations are binary, either the baby is crying
or not. The crying observation does not uniquely identify the baby's state, as
the baby may cry due to hunger or discomfort, which makes this a partially
observable problem. The agent receives a reward when it correctly addresses the
baby's needs and a penalty when the wrong action is taken.
\VI{
\subsubsection{Rock Sample}
In the Rock Sample problem \cite{Smith04uai} a rover explores a $k{\times}k$
grid containing $n$ rocks whose quality (good / bad) is initially unknown; where
$n=15$ and $k=3$ in the experiments section. A state is the rover cell together
with an $n$-bit vector of rock qualities. The agent can move deterministically
in the four cardinal directions, execute a \textit{Sample} action at its current
cell, or invoke one of $n$ sensing actions. The sensing action returns the
observation ${\mathsf{good},\mathsf{bad}}$ with an accuracy level whose
determined as a function of the Manhattan distance to rock $i$; all other
actions yield the null observation. Each step incurs a penalty
$r_{\text{step}}<0$, sensing adds $r_{\text{sensor}}<0$, sampling yields
$r_{\text{good}}>0$ if the rock is good and $r_{\text{bad}}<0$ otherwise, and
reaching the exit column grants $r_{\text{exit}}>0$. These dynamics create a
canonical exploration-exploitation trade-off in which the rover must decide
whether to spend time and energy gathering information or head directly for the
exit.

\subsubsection{Navigate To Goal}
This environment is a $k{\times}n$ grid (default $5{\times}5$) in which a state
is the agent’s coordinate pair $s=(x,y)$. Five actions are
available, \textit{Stay}, \textit{East}, \textit{West}, \textit{North}, and
\textit{South}. An intended move succeeds with probability $0.7$; the remaining
$0.3$ is shared uniformly among the other adjacent cardinal neighbors (attempts
to leave the grid leave the agent in place). Rewards are $+10$ for entering a
goal cell, $-10$ for stepping on the a trap cell, and $0$ elsewhere.
}
\subsection{Hyperparameters}
The hyperparameters for both DB-DESPOT and AR-DESPOT algorithms were selected
through a grid search. We explored an array of parameters for AR-DESPOT,
choosing the highest-performing configuration. Specifically, the hyperparameter
$K$ was varied across $\{10, 50, 500, 5000\}$, while $\lambda$ was evaluated at
$\{0, 0.01, 0.1\}$. Similarly, DB-POMCP and POMCP were examined three different
values for the exploration-exploitation weight, $c=\{0.1, 1.0, 10.0\}$
multiplied by $V_{max}$, which denotes an upper bound for the value function.

For the initialization of the upper and lower bounds used by the algorithms, we
used the maximal reward, multiplied by the remaining time steps of the episode,
$\mathcal{R}_{\max}\cdot (\mathcal{T}-t-1)$.

\end{document}